\documentclass{article} 
\usepackage[T1]{fontenc} 
\usepackage{iclr2027_conference,times}

\usepackage{amsmath,amsfonts,bm}

\def\eqref#1{equation~\ref{#1}}

\def\1{\bm{1}}

\DeclareMathAlphabet{\mathsfit}{\encodingdefault}{\sfdefault}{m}{sl}
\SetMathAlphabet{\mathsfit}{bold}{\encodingdefault}{\sfdefault}{bx}{n}

\usepackage{hyperref}
\hypersetup{
  hidelinks,
  pdftitle={TaxoMFM: A Taxonomy-Aware Foundation Model for the Human Gut Microbiome},
  pdfauthor={Haochen Xue, Anastasia Godneva, Yulong Li, Qi Cao, Zhaodong Wu, Kaiyu Liang, Jionglong Su, Eran Segal, Imran Razzak}
}
\usepackage{url}
\usepackage{graphicx}
\usepackage{booktabs}
\usepackage{longtable}
\usepackage{amssymb}
\usepackage{multirow}
\usepackage{array}
\usepackage{tabularx}
\usepackage{xcolor}
\usepackage{colortbl}
\usepackage{float}
\makeatletter
\newcommand{\captionof}[1]{\def\@captype{#1}\caption}
\makeatother

\title{TaxoMFM: A Taxonomy-Aware Foundation Model for the Human Gut Microbiome}

\author{%
\begin{minipage}[t]{\dimexpr\textwidth-2\tabcolsep\relax}
\centering\normalfont
\textbf{Haochen Xue\textsuperscript{1,2,*}, Anastasia Godneva\textsuperscript{3,*}, Yulong Li\textsuperscript{2,*}, Qi Cao\textsuperscript{1},}\\[2pt]
\textbf{Zhaodong Wu\textsuperscript{1}, Kaiyu Liang\textsuperscript{1}, Jionglong Su\textsuperscript{1}, Eran Segal\textsuperscript{3}, Imran Razzak\textsuperscript{2}}\\[6pt]
{\small
\textsuperscript{1}Xi'an Jiaotong-Liverpool University\\
\textsuperscript{2}Mohamed bin Zayed University of Artificial Intelligence\\
\textsuperscript{3}Weizmann Institute of Science}
\end{minipage}%
}

\definecolor{tablegreen}{HTML}{16883E}
\newcommand{\tablecheck}{\textcolor{tablegreen}{\ensuremath{\checkmark}}}
\newcommand{\tablecheckbold}{\textcolor{tablegreen}{\ensuremath{\boldsymbol{\checkmark}}}}

\iclrfinalcopy 
\begin{document}

\maketitle
\begingroup
\renewcommand{\thefootnote}{\fnsymbol{footnote}}
\footnotetext[1]{These authors contributed equally to this work.}
\endgroup
\fancyhead{}
\renewcommand{\headrulewidth}{0pt}
\raggedbottom 

\begin{abstract}
Existing gut microbiome foundation models typically use flat taxon representations, limiting their ability to exploit known taxonomic hierarchies. We introduce TaxoMFM, a microbiome foundation model that integrates taxonomic structure into self-supervised learning. TaxoMFM jointly learns abundance patterns across taxonomic ranks and relative proportions within groups through taxonomy-aware encoding and hierarchical compositional reconstruction. We further construct TaxoGut-HPP, comprising 100,806 gut microbiome profiles from 231 source cohorts and covering 40 classification and 8 regression tasks. TaxoMFM achieves competitive performance across disease, gut health, host trait, and liver ultrasound phenotype prediction. In cross-cohort zero-shot transfer, where target cohorts are excluded from both pretraining and downstream adaptation, TaxoMFM achieves AUROCs of 0.838, 0.736, and 0.722 for colorectal cancer, type 2 diabetes, and inflammatory bowel disease, respectively. Further analyses show that TaxoMFM recovers community changes across taxonomic scales, while preserving community information and producing consistent representations under database updates and species-vocabulary shifts. \mbox{Our Code is available on
{\hypersetup{pdfborder={0 0 0}}
\href{https://anonymous.4open.science/r/TaxoMFM-F16D/}
{\textcolor[HTML]{147FB3}{\underline{GitHub}}}}.}
\end{abstract}

\section{Introduction}

Foundation models learn general representations from large unlabeled datasets and transfer them to a wide range of downstream tasks \citep{bommasani2021foundation}. This approach is well suited to gut microbiome data. Gut microbiome profiles are high-dimensional and sparse \citep{oh2020deepmicro}. Recent gut microbiome foundation models combine multiple public cohorts for self-supervised pretraining and transfer the learned representations to disease prediction tasks \citep{pope2025microbiomes,medearis2026biomegpt,karwowska2026gutformer}. These studies show the potential of pretraining for microbiome modeling across tasks.

However, microbial signals associated with diseases and host phenotypes can occur at several taxonomic scales. Changes may involve a single species, an entire genus, or a family. For example, a genus can increase in total relative abundance, or maintain the same total while its species proportions change. The taxonomic hierarchy shows how abundant each group is in the community and how its members contribute to that abundance. Current gut microbiome foundation models generally use flat taxon representations that do not explicitly encode these relationships. Previous statistical and machine learning studies have shown that taxonomic or phylogenetic information can help characterize community differences and support downstream prediction \citep{silverman2017philr,reiman2020popphycnn,jiang2025miostone}. We ask how self-supervised pretraining can use this hierarchy to preserve both abundance distributions across taxonomic scales and composition within lineages.

\suppressfloats[t]
\begin{figure}[t]
    \centering
    \includegraphics[width=0.75\linewidth]{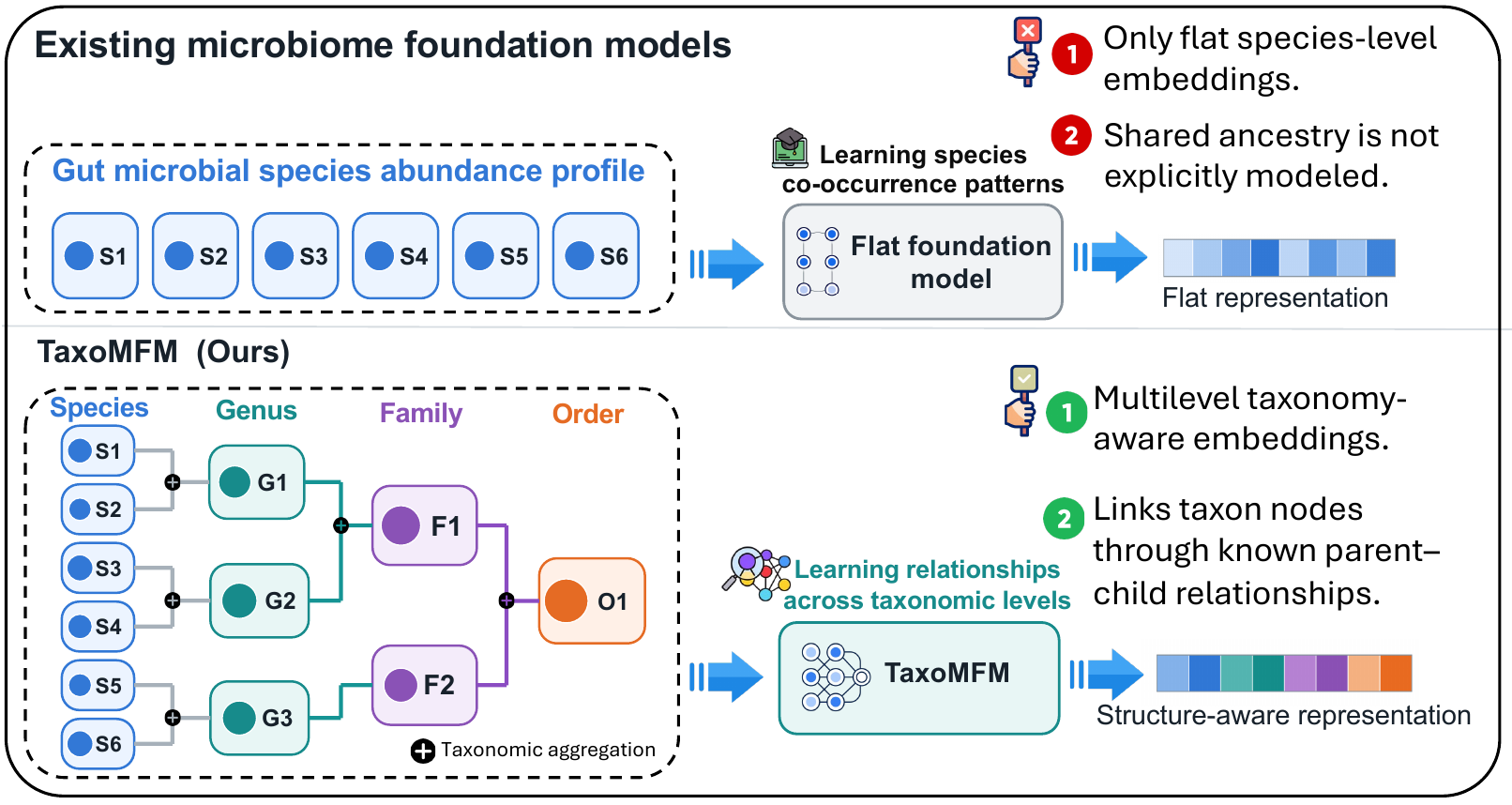}
    \caption{Conceptual comparison of flat and taxonomy-aware microbiome foundation models. Both approaches use the same species-level abundance profile. Existing models represent species as a flat set, whereas TaxoMFM aggregates their abundances along known taxonomic lineages and jointly models species, genus, family, and order nodes to learn a structure-aware representation.}
    \label{fig:concept}
    \vspace{-10pt}
\end{figure}

Existing evaluations largely focus on case--control classification in public disease cohorts. Gut symptoms, such as abdominal pain and bloating, and continuous physiological traits receive less attention. This narrow scope leaves it unclear whether the learned representations capture variation in gut health and host physiology beyond disease status.

To address the modeling limitation, we introduce TaxoMFM, a foundation model that explicitly incorporates the taxonomic hierarchy into both encoding and reconstruction. Its encoder uses relation-biased self-attention to jointly model species, genus, family, and order. Its decoder predicts child proportions within each parent and reconstructs abundances along the taxonomy, ensuring that child abundances sum to their parent's abundance. Pretraining combines multiscale composition reconstruction with hierarchical composition learning, allowing the model to learn both group abundances and the relative proportions of their members. Together, these components provide a way to learn representations from compositional data with a known hierarchy. Fig.~\ref{fig:concept} contrasts this approach with flat species-level pretraining.

To broaden the data and evaluation coverage, we construct TaxoGut-HPP, comprising 100,806 gut microbiome profiles from 231 source cohorts. It covers 40 classification and 8 regression tasks spanning disease, gut health, host traits, and continuous clinical phenotypes. This coverage extends evaluation beyond disease classification to gut symptoms and quantitative physiological measurements. Table~\ref{tab:gut-pretraining-comparison} compares its scale and phenotype coverage with existing pretraining resources. Using TaxoGut-HPP, we evaluate TaxoMFM through downstream prediction, cross-cohort transfer, ablations, and analyses of multiscale recovery and consistency across taxonomic vocabularies.

Our main contributions are as follows:
\begin{enumerate}
    \item We introduce TaxoMFM, combining taxonomy-aware encoding and hierarchical compositional decoding to learn transferable microbiome representations through joint reconstruction of multiscale abundances and within-lineage proportions.
    \item We construct TaxoGut-HPP with 100,806 profiles from 231 source cohorts, expanding pretraining data and evaluation across disease, gut health, host traits, and continuous clinical phenotypes.
    \item TaxoMFM outperforms existing foundation models and conventional machine learning methods across a broad range of downstream tasks. Ablations and structural analyses assess the contributions of taxonomy and hierarchical composition learning.
\end{enumerate}

\vspace{-7pt}
\begin{table}[!b]
    \caption{Comparison of pretraining and downstream evaluation datasets used by human gut microbiome foundation models.}
    \label{tab:gut-pretraining-comparison}
    \centering
    \scriptsize
    \setlength{\tabcolsep}{2pt}
    \renewcommand{\arraystretch}{1.15}
    \resizebox{\textwidth}{!}{%
    \begin{tabular}{l r r r r r c c c c}
        \toprule
        \multirow{2}{*}{\textbf{Model}} & \multirow{2}{*}{\textbf{Size}} & \multirow{2}{*}{\textbf{Cohorts}} & \multicolumn{3}{c}{\textbf{Downstream task counts}} & \multicolumn{4}{c}{\textbf{Downstream task coverage}} \\
        \cmidrule(lr){4-6}\cmidrule(l){7-10}
        & & & \textbf{Total} & \textbf{Classification} & \textbf{Regression} & \textbf{Disease} & \textbf{Gut health} & \textbf{Host traits} & \textbf{Clinical phenotype} \\
        \midrule
        Pope et al.~\citep{pope2025microbiomes}
            & 18,480 & 1 & 3 & 3 & 0 & \tablecheck & $\times$ & \tablecheck & $\times$ \\
        BiomeGPT~\citep{medearis2026biomegpt}
            & 13,524 & 86 & 27 & 27 & 0 & \tablecheck & $\times$ & $\times$ & $\times$ \\
        GUT-FORMer~\citep{karwowska2026gutformer}
            & 21,024 & 93 & 27 & 25 & 2 & \tablecheck & $\times$ & \tablecheck & $\times$ \\
        MBEmbed~\citep{zahavi2025mbembed}
            & 85,364 & 3 & 7 & 4 & 3 & \tablecheck & $\times$ & \tablecheck & \tablecheck \\
        \midrule
        \textbf{TaxoMFM (ours)}
            & \textbf{100,806} & \textbf{231} & \textbf{48} & \textbf{40} & \textbf{8} & \tablecheckbold & \tablecheckbold & \tablecheckbold & \tablecheckbold \\
        \bottomrule
    \end{tabular}%
    }
\end{table}

\section{Related Work}

\subsection{Microbiome foundation models}

Recent studies have applied self-supervised pretraining to microbiome profiles. \citet{pope2025microbiomes} use an ELECTRA-style objective to detect replaced microbial tokens. MGM~\citep{zhang2026mgm} applies causal language modeling to genus-level profiles, while Waypoint~\citep{treloar2026waypoint} studies causal pretraining with both genus- and species-level tokenization. For the human gut microbiome, BiomeGPT~\citep{medearis2026biomegpt} models species abundance profiles, GUT-FORMer~\citep{karwowska2026gutformer} combines taxonomic and functional profiles, and MBEmbed~\citep{zahavi2025mbembed} learns sample representations from large gut cohorts. These models generally treat each taxon as a separate input unit and learn microbial relationships mainly from co-occurrence. They do not explicitly model the known hierarchy among species, genera, families, and orders during pretraining.

\subsection{Taxonomy-aware microbiome learning}

DeepMicro~\citep{oh2020deepmicro} used autoencoders to learn low-dimensional representations from microbial abundance profiles. Other approaches explicitly incorporated taxonomic or phylogenetic relationships. Ph-CNN~\citep{fioravanti2018phcnn} used phylogenetic distances to define convolutional neighborhoods, while PopPhy-CNN~\citep{reiman2020popphycnn} arranged abundances at leaf and ancestral nodes into a tree-based matrix for convolution. TaxoNN~\citep{sharma2020taxonn} grouped OTUs by phylum and combined features learned per group. More recently, DeepPhylo~\citep{wang2024deepphylo} combined phylogenetic distance embeddings with abundance information for unsupervised analysis and supervised prediction. PhyloGCNE~\citep{dong2026phylogcne} represented microbiome samples as phylogenetic graphs with leaf and ancestral nodes, using edge features to guide message passing. In contrast, TaxoMFM uses relation-biased attention to exchange information across taxonomic ranks. Its hierarchical decoder predicts child proportions within each parent, ensuring that reconstructed child abundances sum to their parent's abundance. Pretraining jointly matches multiscale abundance profiles and within-parent compositions to learn reusable sample representations.

\section{TaxoMFM}
\label{sec:method}

\begin{figure*}[t]
    \centering
    \includegraphics[width=\textwidth]{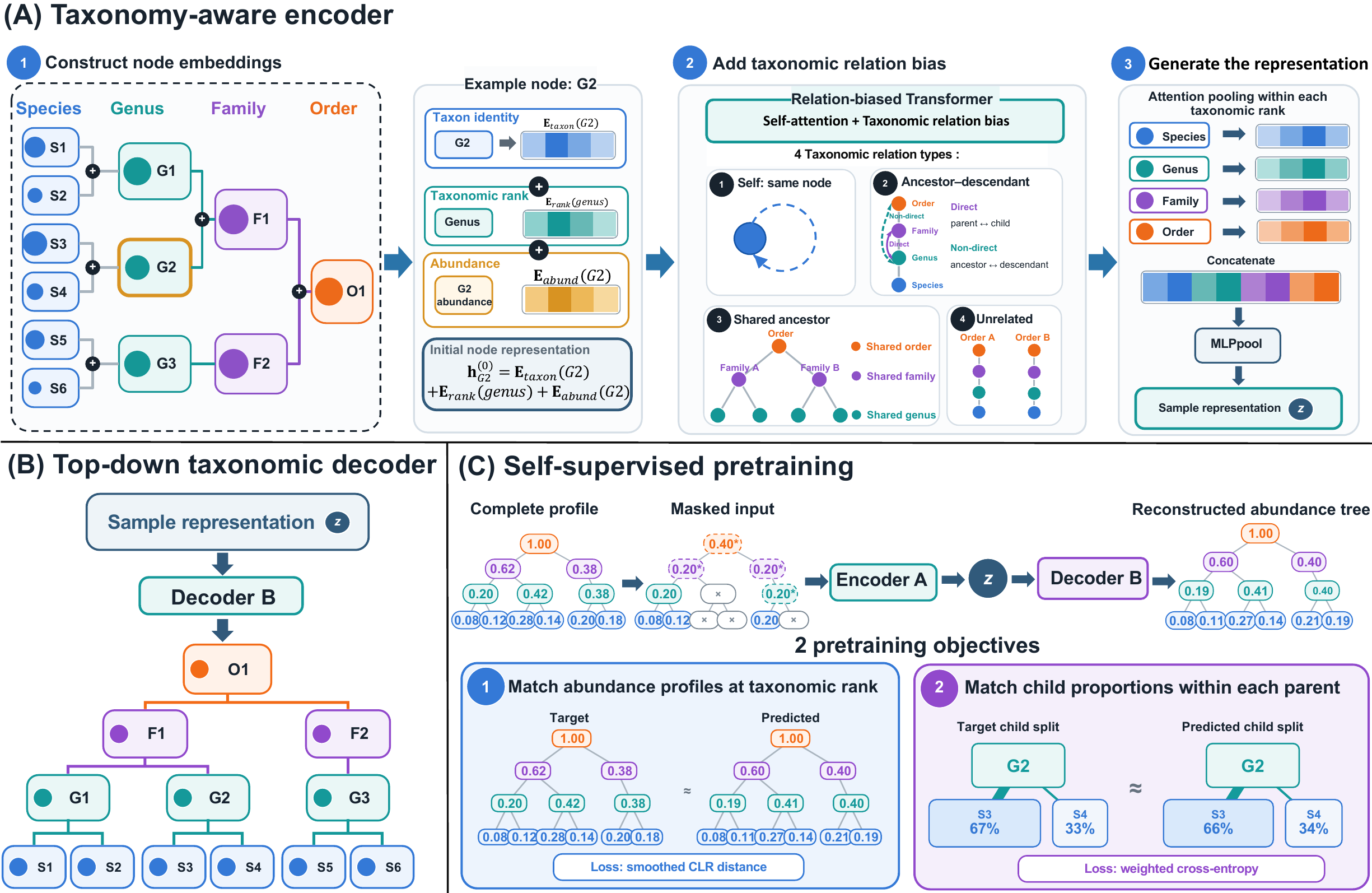}
    \caption{TaxoMFM architecture and self-supervised pretraining. (A) The encoder combines taxon identity, taxonomic rank, and abundance embeddings with relation-biased self-attention and rank-wise pooling to generate the sample representation $z$. (B) The decoder predicts child proportions, reconstructing abundance profiles top-down through the taxonomy. (C) Pretraining masks species or subtrees and optimizes full-tree reconstruction and within-parent child-proportion matching.}
    \label{fig:taxomfm_framework}
\end{figure*}

TaxoMFM consists of a taxonomy-aware encoder and a hierarchical compositional decoder. The encoder uses known taxonomic relations to learn a structure-aware representation of each microbiome sample. During pretraining, species and taxonomic subtrees in the encoder input are masked, and the decoder provides the training signal for the encoder by reconstructing complete abundance profiles through the known hierarchy. After pretraining, the decoder is discarded and the encoder is used for downstream tasks. Fig.~\ref{fig:taxomfm_framework} summarizes the full framework.

\subsection{Taxonomy-aware encoder}
\label{sec:taxonomy_encoder}

TaxoMFM takes a species-level abundance profile together with the taxonomic lineage of each species, which is typically available in existing microbiome datasets. Using these lineages, species assigned to the same ancestor are aggregated as
\begin{equation}
    x_j^{(\ell)}
    =
    \sum_{i:\,\operatorname{ancestor}_{\ell}(i)=j}
    x_i^{(\mathrm{species})},
    \qquad
    \ell\in\{\mathrm{genus},\mathrm{family},\mathrm{order}\}.
    \label{eq:rank_aggregation}
\end{equation}
Here, $\operatorname{ancestor}_{\ell}(i)$ denotes species $i$'s ancestor at rank $\ell$, so higher-rank abundances are computed from the species input. For each sample, the detected species and their ancestors form a sample-specific active tree. The encoder takes the taxa and abundances from all four levels in this tree as input. For each active node $v$, $x_v$ and $\ell_v$ denote its abundance and taxonomic rank, respectively. During pretraining, $s_v$ records whether the node is observed, masked, or partially observed; these states are defined in Section~\ref{sec:self_supervised_pretraining}. We represent the node as
\begin{equation}
    \mathbf{h}_v
    =\mathbf{E}_{\mathrm{taxon}}[v]
    +\mathbf{E}_{\mathrm{rank}}[\ell_v]
    +\operatorname{MLP}(x_v)
    +\mathbf{E}_{\mathrm{state}}[s_v].
    \label{eq:node_input}
\end{equation}
Here, $\mathbf{E}_{\mathrm{taxon}}[v]$ is a learned embedding for taxon $v$ itself, while $\mathbf{E}_{\mathrm{rank}}[\ell_v]$ indicates whether it is a species, genus, family, or order node. The MLP maps its scalar abundance $x_v$ into the hidden space. The state embedding is used only for masked pretraining.

The encoder uses the taxonomy to guide information exchange between active nodes. For every ordered pair $(u,v)$, the taxonomy defines a relation type $r_{uv}\in\mathcal{R}_{\mathrm{rel}}$. We add a learned relation bias to the standard attention score in each head:
\begin{equation}
    a^{(m)}_{uv}
    =\operatorname{Attn}^{(m)}(\mathbf{h}_u,\mathbf{h}_v)
    +b^{(m)}_{r_{uv}}.
    \label{eq:relation_attention}
\end{equation}
The relation type is determined by the positions of $u$ and $v$ in the taxonomy. It distinguishes self, parent, child, ancestor, and descendant pairs. For nodes on different branches, it records whether their lowest common ancestor is a genus, family, or order. Pairs that meet only at the virtual root are treated as unrelated. These relation types are illustrated in Fig.~\ref{fig:taxomfm_framework}A. Each attention head learns a separate bias for every relation type, allowing known taxonomic proximity to guide information exchange in addition to abundance co-occurrence.

After the final encoder layer, each active node has a contextual representation $\mathbf{h}_v$. We first summarize the nodes separately at each taxonomic level:
\[
    \mathbf{p}^{(r)}
    =\operatorname{AttnPool}_{r}
    \left(\{\mathbf{h}_v\}_{v\in\mathcal{V}_r}\right),
    \qquad
    r\in\{\mathrm{species},\mathrm{genus},\mathrm{family},\mathrm{order}\},
\]
where $\mathcal{V}_r$ contains the active nodes at level $r$. This gives one summary vector for each level. We concatenate the four summaries and map them to the sample representation:
\begin{equation}
    \mathbf{z}
    =\operatorname{MLP}_{\mathrm{pool}}
    \left(
    \operatorname{Concat}
    [\mathbf{p}^{(\mathrm{species})},
    \mathbf{p}^{(\mathrm{genus})},
    \mathbf{p}^{(\mathrm{family})},
    \mathbf{p}^{(\mathrm{order})}]
    \right).
    \label{eq:sample_representation}
\end{equation}
$\mathbf{z}$ feeds the decoder during pretraining and the prediction head downstream.


\subsection{Hierarchical compositional decoding for pretraining}
\label{sec:hierarchical_decoder}
\label{sec:self_supervised_pretraining}

During pretraining, we mask individual species or complete taxonomic subtrees in the sample-specific active tree. Each node is assigned one of the three states used in Equation~\ref{eq:node_input}: observed when its complete abundance is available, masked when its abundance is hidden, and partially observed when its subtree contains both visible and masked taxa, so its input abundance contains only the visible part. The encoder maps this masked tree to the sample representation $\mathbf{z}$. The decoder treats the total community abundance as one and decomposes it along the taxonomy. It first predicts the proportion assigned to each order, and then recursively predicts the proportions assigned to child families, genera, and species. Reconstructing the complete profiles in this way requires $\mathbf{z}$ to encode both the total abundance of each lineage and the relative composition among its children.

The decoder follows the compositional factorization defined by a fixed taxonomic vocabulary shared across samples. For a parent $p$ with abundance $x_p>0$ and full child set $\mathcal{C}(p)$ in this vocabulary, the relative abundance of child $c\in\mathcal{C}(p)$ is
\begin{equation}
    q_{c\mid p}=\frac{x_c}{x_p},
    \qquad
    x_c=x_pq_{c\mid p}.
    \label{eq:hierarchical_factorization}
\end{equation}
The parent abundance $x_p$ describes the total abundance of a lineage, while $q_{c\mid p}$ describes its composition among the children. Together, they separate changes in total lineage abundance from changes within the lineage. Given the sample representation $\mathbf{z}$, a shared decoder predicts the conditional composition by scoring each parent--child pair and normalizing across the children:
\begin{equation}
\begin{aligned}
    s_{pc}
    &=g_{\mathrm{dec}}
    \left(
    \mathbf{z},
    \mathbf{E}_{\mathrm{taxon}}[p],
    \mathbf{E}_{\mathrm{taxon}}[c],
    \mathbf{E}_{\mathrm{edge}}[p,c]
    \right),\\
    \hat q_{c\mid p}
    &=\frac{\exp(s_{pc})}
    {\sum_{k\in\mathcal{C}(p)}\exp(s_{pk})}.
\end{aligned}
    \label{eq:conditional_composition}
\end{equation}
The taxon embeddings identify the parent and child, and the edge embedding identifies their taxonomic-level transition. The same scoring function is shared across all levels. Starting from the virtual root, the decoder converts the predicted conditional compositions into node abundances:
\begin{equation}
    \hat x_{\mathrm{root}}=1,
    \qquad
    \hat x_c=\hat x_p\hat q_{c\mid p}.
    \label{eq:top_down_reconstruction}
\end{equation}
Because the predicted proportions of a parent's children sum to one, their reconstructed abundances also sum to $\hat x_p$. The decoder therefore produces consistent profiles across all four levels without an additional hierarchy penalty. We use the hierarchical decoder to pretrain the encoder with two objectives: multiscale composition reconstruction and hierarchical composition learning.

\noindent\textbf{Multiscale composition reconstruction.}
The decoder reconstructs the complete community profile over the full vocabulary at each taxonomic level. For each rank $r$, we compare the predicted profile $\hat{\mathbf{x}}^{(r)}$ with the corresponding uncorrupted target $\mathbf{x}^{(r)}$, whose entries are zero for undetected taxa. Because microbial abundances are compositional, we use a smoothed centered log-ratio (CLR) distance~\citep{aitchison1986statistical}:
\begin{equation}
    \mathcal{L}_{\mathrm{rec}}
    =\sum_{r\in\{\mathrm{species},\mathrm{genus},\mathrm{family},\mathrm{order}\}}
    \alpha_r
    D_{\mathrm{CLR}}
    \left(\mathbf{x}^{(r)},\hat{\mathbf{x}}^{(r)}\right).
    \label{eq:reconstruction_loss}
\end{equation}
Here, $D_{\mathrm{CLR}}$ is the mean squared distance between the smoothed CLR transformations of two profiles, and $\alpha_r$ balances the contributions of different taxonomic levels. This objective requires $\mathbf{z}$ to preserve how community abundance is distributed across lineages at multiple biological scales.

\noindent\textbf{Hierarchical composition learning.}
Matching the abundance profiles at each taxonomic level does not explicitly emphasize the composition within individual lineages. We therefore supervise the predicted child distribution within every parent:
\begin{equation}
    \mathcal{L}_{\mathrm{hier}}
    =-
    \sum_{p\in\mathcal{P}:x_p>0}
    \omega_p
    \sum_{c\in\mathcal{C}(p)}
    q_{c\mid p}\log\hat q_{c\mid p},
    \label{eq:hierarchical_loss}
\end{equation}
Here, $\mathcal{P}$ contains the virtual root and all order, family, and genus parents. We set $\omega_p$ proportional to $\sqrt{x_p}$ and normalize the weights within each parent rank. This gives more weight to abundant lineages without allowing them to dominate the loss. By directly supervising each parent--child transition, this objective captures compositional rearrangements among related taxa, including species replacement within a genus. The full objective is
\begin{equation}
    \mathcal{L}
    =\lambda_{\mathrm{rec}}\mathcal{L}_{\mathrm{rec}}
    +\lambda_{\mathrm{hier}}\mathcal{L}_{\mathrm{hier}}.
    \label{eq:total_pretraining_loss}
\end{equation}
Multiscale reconstruction preserves the community profile across taxonomic levels, whereas hierarchical composition learning explicitly preserves the local composition within each lineage. Both losses are backpropagated through the hierarchical decoder to train the encoder. After pretraining, the decoder is discarded, and downstream tasks use the taxonomy-aware encoder and the resulting sample representation $\mathbf{z}$.

\ifdefined\experimentappendix

The main classification table reports 12 representative tasks, and Tables~\ref{tab:appendix_general_immune}--\ref{tab:appendix_hpp_classification_continued} provide 28 additional classification tasks. The main regression table reports six representative tasks, and Table~\ref{tab:appendix_hpp_regression} provides two additional HPP regression tasks. Together, these tables cover 40 classification and eight regression tasks. Binary classification tasks are evaluated using AUROC. For the three ordinal multiclass outcomes, abdominal-pain frequency, soft-stool frequency, and hard-stool frequency, we report macro-averaged one-vs-rest AUROC. Regression tasks are evaluated using Spearman's $\rho$ and $R^2$.

\begingroup
\definecolor{appendixresultgreen}{RGB}{0,104,99}
\definecolor{appendixgroupgray}{RGB}{78,78,78}
\newcommand{\appendixgroup}[2]{%
  \specialrule{0.35pt}{2.2pt}{0pt}
  \rowcolor{black!7}\multicolumn{#1}{@{}l}{\textcolor{appendixgroupgray}{\textbf{#2}}}\\
  \addlinespace[0.5pt]
}
\newcommand{\appendixheadfive}[1]{\parbox[c]{0.78in}{\centering\bfseries #1}}
\newcommand{\appendixheadsix}[1]{\parbox[c]{0.65in}{\centering\bfseries #1}}
\newcommand{\appendixheadseven}[1]{\parbox[c]{0.56in}{\centering\bfseries #1}}
\newcommand{\appendixheadfour}[1]{\parbox[c]{0.95in}{\centering\bfseries #1}}

\begin{table}[H]
\caption{Additional AUROC results for general and immune-related diseases (part I).}
\label{tab:appendix_general_immune}
\centering
\fontsize{7.6}{8.6}\selectfont
\setlength{\tabcolsep}{2.0pt}
\renewcommand{\arraystretch}{1.06}
\begin{tabular*}{\textwidth}{@{\extracolsep{\fill}}l*{5}{c}@{}}
\toprule
\textbf{Model} &
\appendixheadfive{Healthy vs. diseased} &
\appendixheadfive{Crohn's disease} &
\appendixheadfive{Ulcerative colitis} &
\appendixheadfive{Multiple sclerosis} &
\appendixheadfive{Rheumatoid arthritis}\\
\midrule
\appendixgroup{6}{Standard baselines}
Elastic Net &0.760 &0.872&0.849&0.869&0.834\\
Random Forest & 0.782 &0.908&0.872&0.901&0.859\\
DeepMicro & 0.809 &0.873&0.868&0.908&0.861\\
ExtraTrees & 0.773 &0.902&0.882&0.895&0.864\\
\appendixgroup{6}{Taxonomy-aware models}
Ph-CNN & 0.799 & 0.896 & 0.884 & 0.910& 0.868\\
PopPhy-CNN  & 0.811 &0.907 & 0.899 & 0.923& 0.883\\
MDeep & 0.846 &0.933&0.793&0.943&0.910\\
DeepPhylo & 0.802 & 0.901& 0.878 & 0.918& 0.877\\
DeepBiome & 0.816 & 0.914& 0.903 & 0.951 & 0.891\\
PhyloGCNE &0.822  & 0.919 &0.910 & 0.931&0.903\\
Phylo-Spec & 0.784 &0.878&0.847&0.883&0.859\\
MIOSTONE & 0.838 &0.883&0.927&0.934&0.936\\
\appendixgroup{6}{General foundation models}
MGM & 0.858 &0.948&0.934 &0.947&0.931\\
Waypoint-45M & 0.842 &0.936&0.936&0.950&0.933\\
\appendixgroup{6}{Gut foundation models}
Pope et al. & 0.851 &0.953 & 0.945& 0.951& 0.938\\
BiomeGPT & 0.874 &\underline{0.964}&0.951&\underline{0.963}&\underline{0.959}\\
GUT-FORMer & \underline{0.879} & 0.955& \underline{0.959}& 0.959& 0.958\\
MBEmbed & 0.869 & 0.947 & 0.944& 0.949& 0.946\\
\specialrule{0.75pt}{2pt}{1pt}
\rowcolor{appendixresultgreen!8}\textcolor{appendixresultgreen}{\textbf{TaxoMFM (ours)}} &
\textbf{0.891}&\textbf{0.979}&\textbf{0.967}&\textbf{0.976}&\textbf{0.971}\\
\bottomrule
\end{tabular*}

\end{table}

\clearpage
\begin{table}[H]
\caption{Additional AUROC results for general and immune-related diseases (part II).}
\label{tab:appendix_general_immune_continued}
\centering
\fontsize{7.6}{8.6}\selectfont
\setlength{\tabcolsep}{2.5pt}
\renewcommand{\arraystretch}{1.06}
\begin{tabular*}{\textwidth}{@{\extracolsep{\fill}}l*{4}{c}@{}}
\toprule
\textbf{Model} &
\appendixheadfour{Ankylosing spondylitis} &
\appendixheadfour{Graves' disease} &
\appendixheadfour{Beh\c{c}et's disease} &
\appendixheadfour{Myalgic encephalomyelitis / chronic fatigue syndrome}\\
\midrule
\appendixgroup{5}{Standard baselines}
Elastic Net & 0.892&0.883&0.904&0.862\\
Random Forest & 0.923&0.902&0.947&0.889\\
DeepMicro & 0.907&0.910&0.922&0.902\\
ExtraTrees & 0.917&0.930&0.941&0.877\\
\appendixgroup{5}{Taxonomy-aware models}
Ph-CNN & 0.921& 0.928& 0.935& 0.904\\
PopPhy-CNN & 0.930& 0.941& 0.954& 0.917\\
MDeep & 0.947&0.951&0.961&0.953\\
DeepPhylo & 0.941& 0.936& 0.937& 0.912\\
DeepBiome & 0.938& 0.947& 0.949& 0.927\\
PhyloGCNE & 0.933&0.949&0.938&0.920\\
Phylo-Spec & 0.902&0.908&0.917&0.886\\
MIOSTONE & 0.945&0.955&0.972&0.932\\
\appendixgroup{5}{General foundation models}
MGM & 0.964&0.977&0.976&0.939\\
Waypoint-45M & 0.960&0.963&0.966&0.947\\
\appendixgroup{5}{Gut foundation models}
Pope et al. & 0.975& 0.970& \underline{0.993}& 0.938\\
BiomeGPT & \underline{0.983}&0.982&0.983&0.960\\
GUT-FORMer & 0.979& \underline{0.986}& 0.991& \underline{0.969}\\
MBEmbed & 0.970& 0.968& 0.971& 0.944\\
\specialrule{0.75pt}{2pt}{1pt}
\rowcolor{appendixresultgreen!8}\textcolor{appendixresultgreen}{\textbf{TaxoMFM (ours)}} &
\textbf{0.992}&\textbf{0.993}&\textbf{0.996}&\textbf{0.978}\\
\bottomrule
\end{tabular*}

\end{table}

\begin{table}[H]
\caption{Additional AUROC results for neoplastic and cardiometabolic diseases.}
\label{tab:appendix_neoplastic_metabolic}
\centering
\fontsize{7.6}{8.6}\selectfont
\setlength{\tabcolsep}{1.5pt}
\renewcommand{\arraystretch}{1.06}
\begin{tabular*}{\textwidth}{@{\extracolsep{\fill}}l*{6}{c}@{}}
\toprule
\textbf{Model} &
\appendixheadsix{Colorectal adenoma} &
\appendixheadsix{Pancreatic cancer} &
\appendixheadsix{Coronary artery disease} &
\appendixheadsix{Athero\-sclerotic cardio\-vascular disease} &
\appendixheadsix{Impaired glucose tolerance} &
\appendixheadsix{Melanoma}\\
\midrule
\appendixgroup{7}{Standard baselines}
Elastic Net & 0.833&0.896&0.874&0.903&0.788&0.874\\
Random Forest & 0.868&0.932&0.903&0.934&0.817&0.906\\
DeepMicro & 0.851&0.918&0.882&0.919&0.825&0.895\\
ExtraTrees & 0.873&0.925&0.896&0.928&0.806&0.899\\
\appendixgroup{7}{Taxonomy-aware models}
Ph-CNN & 0.880& 0.939& 0.903& 0.929& 0.823& 0.902\\
PopPhy-CNN & 0.892& 0.944& 0.921& 0.947& 0.843& 0.916\\
MDeep & 0.915&0.952&0.945&0.955&0.864&0.924\\
DeepPhylo & 0.873& 0.936& 0.914& 0.936& 0.887& 0.913\\
DeepBiome & 0.887& 0.948& 0.930& 0.949& 0.856& 0.939\\
PhyloGCNE & 0.899&0.940&0.926&0.941&0.849&0.930\\
Phylo-Spec & 0.858&0.922&0.889&0.919&0.814&0.877\\
MIOSTONE & 0.909&0.954&0.938&0.953&0.861&0.933\\
\appendixgroup{7}{General foundation models}
MGM & 0.928&0.962&0.943&0.976&0.872&0.941\\
Waypoint-45M & 0.919&0.968&0.934&0.962&0.881&0.940\\
\appendixgroup{7}{Gut foundation models}
Pope et al. & \underline{0.949}& 0.974& 0.952& 0.975& 0.881& 0.948\\
BiomeGPT & 0.931&\underline{0.982}&\underline{0.962}&\underline{0.988}&\underline{0.899}&0.962\\
GUT-FORMer & 0.940& 0.979& 0.956& 0.979& 0.887& 0.954\\
MBEmbed & 0.947& 0.963& 0.944& 0.970& 0.877& \underline{0.968}\\
\specialrule{0.75pt}{2pt}{1pt}
\rowcolor{appendixresultgreen!8}\textcolor{appendixresultgreen}{\textbf{TaxoMFM (ours)}} &
\textbf{0.962}&\textbf{0.993}&\textbf{0.975}&\textbf{0.995}&\textbf{0.919}&\textbf{0.980}\\
\bottomrule
\end{tabular*}

\end{table}

\clearpage
\begin{table}[H]
\caption{Additional AUROC results for neurological, infectious, and host-related tasks.}
\label{tab:appendix_neuro_infectious}
\centering
\fontsize{7.6}{8.6}\selectfont
\setlength{\tabcolsep}{1.5pt}
\renewcommand{\arraystretch}{1.06}
\begin{tabular*}{\textwidth}{@{\extracolsep{\fill}}l*{6}{c}@{}}
\toprule
\textbf{Model} &
\appendixheadsix{Parkinson's disease} &
\appendixheadsix{Schizo\-phrenia} &
\appendixheadsix{Soil-trans\-mitted helminths} &
\appendixheadsix{\textit{Clostridi\-oides difficile} infection} &
\appendixheadsix{Acute diarrhea} &
\appendixheadsix{Premature birth}\\
\midrule
\appendixgroup{7}{Standard baselines}
Elastic Net & 0.898&0.883&0.916&0.869&0.922&0.904\\
Random Forest & 0.924&0.918&0.944&0.899&0.957&0.932\\
DeepMicro & 0.916&0.925&0.930&0.920&0.949&0.941\\
ExtraTrees & 0.919&0.911&0.956&0.883&0.950&0.928\\
\appendixgroup{7}{Taxonomy-aware models}
Ph-CNN & 0.931& 0.928& 0.950& 0.916& 0.959& 0.937\\
PopPhy-CNN & 0.941& 0.948& 0.964& 0.934& 0.969& 0.949\\
MDeep & 0.953&0.959&0.969&0.949&0.973&0.962\\
DeepPhylo & 0.932& 0.946& 0.947& 0.923& 0.958& 0.941\\
DeepBiome & 0.944& 0.951& 0.960& 0.938& 0.965& 0.953\\
PhyloGCNE & 0.946&0.941&0.962&0.940&0.978&0.951\\
Phylo-Spec & 0.913&0.910&0.939&0.907&0.931&0.918\\
MIOSTONE & 0.951&0.964&0.966&0.945&0.971&0.959\\
\appendixgroup{7}{General foundation models}
MGM & 0.971&0.972&0.978&0.955&0.977&0.978\\
Waypoint-45M & 0.965&0.966&0.987&0.951&0.973&0.971\\
\appendixgroup{7}{Gut foundation models}
Pope et al. & 0.979& 0.973& 0.985& \textbf{0.980}& 0.986& 0.974\\
BiomeGPT & \underline{0.990}&\underline{0.985}&\underline{0.992}&0.969&0.989&\underline{0.989}\\
GUT-FORMer & 0.985& 0.979& 0.982& \underline{0.978}& \underline{0.993}& 0.981\\
MBEmbed & 0.973& 0.981& 0.988& 0.959& 0.979& 0.986\\
\specialrule{0.75pt}{2pt}{1pt}
\rowcolor{appendixresultgreen!8}\textcolor{appendixresultgreen}{\textbf{TaxoMFM (ours)}} &
\textbf{0.998}&\textbf{0.996}&\textbf{0.993}&\underline{0.978}&\textbf{0.996}&\textbf{0.998}\\
\bottomrule
\end{tabular*}

\end{table}

\begin{table}[H]
\caption{Additional HPP classification results for gastrointestinal outcomes. Soft- and hard-stool frequency use macro-AUROC; the remaining tasks use AUROC.}
\label{tab:appendix_hpp_classification}
\centering
\fontsize{7.6}{8.6}\selectfont
\setlength{\tabcolsep}{2.0pt}
\renewcommand{\arraystretch}{1.06}
\begin{tabular*}{\textwidth}{@{\extracolsep{\fill}}l*{5}{c}@{}}
\toprule
\textbf{Model} &
\appendixheadfive{Soft-stool frequency} &
\appendixheadfive{Hard-stool frequency} &
\appendixheadfive{Nausea in the past three months} &
\appendixheadfive{Anorectal disease} &
\appendixheadfive{Gallstone / cholecystectomy}\\
\midrule
\appendixgroup{6}{Standard baselines}
Elastic Net & 0.596&0.587&0.664&0.624&0.669\\
Random Forest & 0.638& 0.625& 0.698& 0.657& 0.706\\
DeepMicro & 0.632& 0.619& 0.728& 0.668& 0.719\\
ExtraTrees & 0.613& 0.621& 0.687& 0.646& 0.698\\
\appendixgroup{6}{Taxonomy-aware models}
Ph-CNN & 0.639& 0.621& 0.718& 0.668& 0.726\\
PopPhy-CNN & 0.651& 0.640& 0.734& 0.687& 0.742\\
MDeep & 0.677& 0.668& 0.754& 0.728& 0.789\\
DeepPhylo & 0.648& 0.637& 0.730& 0.704& 0.745\\
DeepBiome & 0.678& 0.667& 0.756& 0.709& 0.768\\
PhyloGCNE & 0.669& 0.660& 0.768& 0.714& 0.764\\
Phylo-Spec & 0.631& 0.609& 0.708& 0.660& 0.711\\
MIOSTONE & 0.681& 0.674& 0.781& 0.741& 0.779\\
\appendixgroup{6}{General foundation models}
MGM & 0.688& 0.677& 0.806& 0.756& 0.819\\
Waypoint-45M & 0.674& 0.689& 0.813& 0.748& 0.810\\
\appendixgroup{6}{Gut foundation models}
Pope et al. & 0.710& 0.704& 0.841& 0.785& 0.831\\
BiomeGPT & 0.724& 0.706& \underline{0.859}& 0.787& 0.849\\
GUT-FORMer & \textbf{0.740}& 0.714& 0.846& \underline{0.796}& 0.833\\
MBEmbed & 0.694& \underline{0.725}& 0.835& 0.778& \underline{0.865}\\
\specialrule{0.75pt}{2pt}{1pt}
\rowcolor{appendixresultgreen!8}\textcolor{appendixresultgreen}{\textbf{TaxoMFM (ours)}} &
\underline{0.738}&\textbf{0.742}&\textbf{0.885}&\textbf{0.810}&\textbf{0.872}\\
\bottomrule
\end{tabular*}
\end{table}

\clearpage
\providecommand{\appendixregpair}[2]{#1\,/\,#2}
\begin{table}[H]
\caption{Additional HPP classification and regression results. The first two columns report AUROC; the final two report Spearman $\rho$ / $R^2$ (all $\uparrow$).}
\label{tab:appendix_hpp_classification_continued}
\label{tab:appendix_hpp_regression}
\centering
\fontsize{7.6}{8.6}\selectfont
\setlength{\tabcolsep}{2.5pt}
\renewcommand{\arraystretch}{1.06}
\begin{tabular*}{\textwidth}{@{\extracolsep{\fill}}l*{4}{c}@{}}
\toprule
\textbf{Model} &
\appendixheadfour{Fatty liver history} &
\appendixheadfour{Peptic ulcer disease} &
\appendixheadfour{Bowel-movement satisfaction} &
\appendixheadfour{Daily bowel-movement count}\\
\midrule
\appendixgroup{5}{Standard baselines}
Elastic Net & 0.657 & 0.629 & \appendixregpair{0.118}{-0.012} & \appendixregpair{0.183}{0.041}\\
Random Forest & 0.741 & 0.612 & \appendixregpair{0.156}{0.019} & \appendixregpair{0.258}{0.096}\\
ExtraTrees & 0.726 & 0.624 & \appendixregpair{0.147}{0.023} & \appendixregpair{0.246}{0.101}\\
DeepMicro & 0.693 & 0.655 & \appendixregpair{0.193}{0.031} & \appendixregpair{0.229}{0.071}\\
\appendixgroup{5}{Taxonomy-aware models}
Ph-CNN & 0.711 & 0.642 & \appendixregpair{0.154}{0.012} & \appendixregpair{0.216}{0.061}\\
PopPhy-CNN & 0.734 & 0.679 & \appendixregpair{0.181}{0.026} & \appendixregpair{0.247}{0.089}\\
MDeep & 0.764 & 0.657 & \appendixregpair{0.211}{0.042} & \appendixregpair{0.281}{0.120}\\
DeepPhylo & 0.722 & 0.705 & \appendixregpair{0.169}{0.021} & \appendixregpair{0.295}{0.141}\\
DeepBiome & 0.751 & 0.671 & \appendixregpair{0.222}{0.046} & \appendixregpair{0.268}{0.109}\\
PhyloGCNE & 0.779 & 0.698 & \appendixregpair{0.105}{0.004} & \appendixregpair{0.255}{0.092}\\
Phylo-Spec & 0.701 & 0.633 & \appendixregpair{0.128}{-0.007} & \appendixregpair{0.237}{0.079}\\
MIOSTONE & 0.803 & 0.691 & \appendixregpair{0.216}{0.048} & \appendixregpair{0.312}{0.156}\\
\appendixgroup{5}{General foundation models}
MGM & 0.808 & 0.703 & \appendixregpair{0.239}{0.058} & \appendixregpair{0.306}{0.162}\\
Waypoint-45M & 0.792 & 0.724 & \appendixregpair{0.228}{0.061} & \appendixregpair{0.324}{0.168}\\
\appendixgroup{5}{Gut foundation models}
Pope et al. & 0.799 & 0.741 & \appendixregpair{0.234}{0.054} & \appendixregpair{0.331}{0.179}\\
BiomeGPT & 0.826 & 0.716 & \appendixregpair{0.253}{0.065} & \appendixregpair{0.317}{0.159}\\
GUT-FORMer & \underline{0.839} & 0.752 & \appendixregpair{0.247}{\underline{0.068}} & \appendixregpair{0.341}{\underline{0.193}}\\
MBEmbed & 0.817 & \underline{0.761} & \appendixregpair{\underline{0.260}}{0.062} & \appendixregpair{\underline{0.352}}{0.186}\\
\specialrule{0.75pt}{2pt}{1pt}
\rowcolor{appendixresultgreen!8}\textcolor{appendixresultgreen}{\textbf{TaxoMFM (ours)}} &
\textbf{0.853} & \textbf{0.776} &
\appendixregpair{\textbf{0.266}}{\textbf{0.072}} &
\appendixregpair{\textbf{0.369}}{\textbf{0.210}}\\
\bottomrule
\end{tabular*}

\end{table}
\endgroup

\else

\section{Experiments}

\subsection{Experimental setting}

Experiments use the full TaxoGut-HPP dataset, which combines a large pretraining corpus with broad downstream coverage. Standard and taxonomy-aware baselines do not use corpus-level pretraining and are trained separately for each downstream task. All experiments use the same task definitions, participant-level splits, and evaluation metrics. In all result tables, bold and underlined values indicate the best and second-best results, respectively.

\vspace{-7pt}

\subsection{Comparative experiments}

\begingroup
\definecolor{resultgreen}{RGB}{0,104,99}
\definecolor{resultgroupgray}{RGB}{78,78,78}
\newcommand{\resultpair}[2]{#1\,/\,#2}
\newcommand{\resultgroup}[2]{%
  \specialrule{0.35pt}{2.2pt}{0pt}
  \rowcolor{black!7}\multicolumn{#1}{@{}l}{\textcolor{resultgroupgray}{\textbf{#2}}}\\
  \addlinespace[0.5pt]
}
\vspace{-20pt}

\begin{table}[H]
\caption{Representative classification performance. Binary tasks use AUROC and abdominal pain frequency uses macro-AUROC ($\uparrow$). The baselines use the locked TaxoGut-HPP tasks and splits.}
\label{tab:classification-results}
\centering
\fontsize{7.35}{8.2}\selectfont
\setlength{\tabcolsep}{1.0pt}
\renewcommand{\arraystretch}{1.05}
\resizebox{\textwidth}{!}{%
\begin{tabular}{@{}l*{5}{c}@{\hspace{4pt}}*{7}{c}@{}}
\toprule
\multirow{2}{*}{\textbf{Model}} &
\multicolumn{5}{c}{\textbf{Disease}} &
\multicolumn{7}{c}{\textbf{Gut health (HPP)}}\\
\cmidrule(lr){2-6}\cmidrule(l){7-13}
& \shortstack{\textbf{Colorectal}\\\textbf{cancer}} &
\shortstack{\textbf{Type 2}\\\textbf{diabetes}} &
\textbf{IBD} &
\shortstack{\textbf{Liver}\\\textbf{cirrhosis}} &
\textbf{NAFLD} &
\shortstack{\textbf{Current}\\\textbf{abdominal pain}} &
\shortstack{\textbf{Abdominal}\\\textbf{pain frequency}} &
\shortstack{\textbf{Current}\\\textbf{bloating}} &
\shortstack{\textbf{GI}\\\textbf{sensitivity}} &
\shortstack{\textbf{Irritable bowel}\\\textbf{syndrome}} &
\shortstack{\textbf{\textit{H. pylori}}\\\textbf{infection/history}} &
\shortstack{\textbf{Gastro-oesophageal}\\\textbf{reflux disease}}\\
\midrule
\resultgroup{13}{Standard baselines}
Elastic Net & 0.826&0.791&0.868&0.811&0.807&0.593&0.572 & 0.637&0.632 & 0.582 &0.628 & 0.651 \\
Random Forest & 0.839&\underline{0.917}&0.853&0.836&0.819&0.607& 0.583 & 0.691&0.646 & 0.547 &0.667 & 0.663 \\
DeepMicro & 0.803&0.763&0.865&0.816& 0.832 &0.602 & 0.574 & 0.658 & 0.627 & 0.562 & 0.628 & 0.646 \\
\resultgroup{13}{Taxonomy-aware models}
Ph-CNN & 0.844 & 0.817 & 0.873 &0.855 & 0.840 & 0.619 & 0.561 & 0.649 & 0.624 & 0.578 & 0.623 & 0.619\\
PopPhy-CNN & 0.840&0.826&0.877 &0.847& 0.838&0.630&0.578&0.673&0.642&0.585& 0.651 & 0.667\\
MDeep & 0.851&0.844&0.901&0.910&0.903&0.682&0.587&0.750&0.664&0.599&0.684&0.703\\
DeepPhylo & 0.833&0.811&0.858 & 0.829& 0.841& 0.622& 0.588 & 0.689& 0.670 & 0.624&0.611& 0.655\\
DeepBiome & 0.836&0.808&0.867&0.845&0.833&0.617 &0.578&0.657&0.669&0.654&0.601&0.642\\
PhyloGCNE & 0.846&0.801&0.883&0.871&0.877&0.637&0.593&0.741&0.679&0.577&0.620&0.688\\
Phylo-Spec & 0.831&0.839&0.860&0.822&0.823&0.609&0.548&0.647&0.653&0.609&0.690&0.675\\
MIOSTONE & 0.855&0.834&0.902&0.860&0.878&0.653& 0.589 &0.724&0.690&0.611&0.698&0.693\\

\resultgroup{13}{General foundation models}
MGM & 0.898&0.864&0.922&0.961&0.947&0.707& 0.632 &0.752 & 0.669 &0.690 &0.752&0.712\\
Waypoint-45M & 0.861&0.872&0.891&0.972&0.935&0.696&0.619 & 0.741 &0.698&0.675&0.711&0.732\\

\resultgroup{13}{Gut foundation models}
Pope et al. & 0.913 & 0.903& 0.937& 0.939& \underline{0.963}&0.711&0.641 &0.751 &0.724 &0.688& 0.768 &0.692\\
BiomeGPT & \underline{0.933}&\underline{0.917}&0.947&\underline{0.984}&0.945&0.732&0.648&0.764&\underline{0.748}&0.728&0.773&0.716\\
GUT-FORMer & 0.928&0.898&\underline{0.966}&0.971&0.952&\underline{0.741}& \underline{0.652} & \underline{0.789}& 0.734 & \underline{0.736} & \underline{0.797} &0.728 \\
MBEmbed & 0.909 & 0.911 & 0.928 & 0.944 & 0.940 & 0.721 & 0.619 &  0.738 &  0.740 &  0.695 &  0.747 &  \underline{0.749} \\
\specialrule{0.75pt}{2pt}{1pt}
\rowcolor{resultgreen!8}\textcolor{resultgreen}{\textbf{TaxoMFM (ours)}} &\textbf{0.953}&\textbf{0.934}&\textbf{0.985}&\textbf{0.996}&\textbf{0.986}&\textbf{0.776}&\textbf{0.665}&\textbf{0.803}&\textbf{0.794}&\textbf{0.745}&\textbf{0.813}&\textbf{0.785}\\

\bottomrule
\end{tabular}%
}
\end{table}

\vspace{-16pt}

\begin{table}[H]
\caption{Representative regression results. In each paired result, the first value is Spearman $\rho$ and the second is $R^2$ ($\uparrow$). The baselines use the locked TaxoGut-HPP tasks and splits.}
\label{tab:regression-results}
\centering
\fontsize{8.0}{9.0}\selectfont
\setlength{\tabcolsep}{2.0pt}
\renewcommand{\arraystretch}{1.08}
\begin{tabular*}{\textwidth}{@{\extracolsep{\fill}}l*{3}{c}@{\hspace{6pt}}*{3}{c}}
\toprule
\multirow{2}{*}{\textbf{Model}} &
\multicolumn{3}{c}{\textbf{Host traits}} &
\multicolumn{3}{c}{\textbf{Liver ultrasound phenotypes}}\\
\cmidrule(lr){2-4}\cmidrule(l){5-7}
& \textbf{Age} & \textbf{BMI} & \shortstack{\textbf{VAT}\\\textbf{mass}} &
\shortstack{\textbf{Attenuation}\\\textbf{coefficient}} &
\shortstack{\textbf{Speed of}\\\textbf{sound}} &
\textbf{Velocity}\\
\midrule
\resultgroup{7}{Standard baselines}
Elastic Net & \resultpair{0.132}{0.038}&\resultpair{0.148}{0.022}&\resultpair{0.214}{0.048}&\resultpair{0.136}{0.038}&\resultpair{0.113}{0.036}&\resultpair{0.111}{0.026}\\
Random Forest & \resultpair{0.117}{0.031}&\resultpair{0.191}{0.045}&\resultpair{0.246}{0.075}&\resultpair{0.128}{0.040}&\resultpair{0.193}{0.042}&\resultpair{0.136}{0.024}\\
DeepMicro & \resultpair{0.280}{0.066}&\resultpair{0.221}{0.065}&\resultpair{0.326}{0.108}&\resultpair{0.162}{0.061}&\resultpair{0.128}{0.033}&\resultpair{0.135}{0.037}\\
\resultgroup{7}{Taxonomy-aware models}
Ph-CNN-Reg & \resultpair{0.178}{0.049}&\resultpair{0.134}{0.048}&\resultpair{0.217}{0.050}&\resultpair{0.139}{0.043}&\resultpair{0.146}{0.027}&\resultpair{0.110}{0.025}\\
PopPhy-CNN-Reg & \resultpair{0.184}{0.028}&\resultpair{0.213}{0.043}&\resultpair{0.238}{0.052}&\resultpair{0.119}{0.026}&\resultpair{0.164}{0.036}&\resultpair{0.125}{0.031}\\
MDeep & \resultpair{0.304}{0.105}&\resultpair{0.246}{0.059}&\resultpair{0.298}{0.113}&\resultpair{0.254}{0.092}&\resultpair{0.132}{0.046}&\resultpair{0.189}{0.051}\\
DeepPhylo & \resultpair{0.212}{0.046}&\resultpair{0.201}{0.064}&\resultpair{0.235}{0.069}&\resultpair{0.145}{0.039}&\resultpair{0.229}{0.074}&\resultpair{0.112}{0.036}\\
DeepBiome & \resultpair{0.288}{0.094}&\resultpair{0.282}{0.087}&\resultpair{0.249}{0.088}&\resultpair{0.153}{0.047}&\resultpair{0.146}{0.048}&\resultpair{0.171}{0.043}\\
PhyloGCNE & \resultpair{0.101}{0.044}&\resultpair{0.118}{0.029}&\resultpair{0.244}{0.083}&\resultpair{0.133}{0.040}&\resultpair{0.155}{0.056}&\resultpair{0.118}{0.032}\\
Phylo-Spec & \resultpair{0.142}{0.052}&\resultpair{0.109}{0.043}&\resultpair{0.207}{0.073}&\resultpair{0.149}{0.058}&\resultpair{0.187}{0.065}&\resultpair{0.115}{0.026}\\
MIOSTONE & \resultpair{0.251}{0.085}&\resultpair{0.239}{0.049}&\resultpair{0.324}{0.125}&\resultpair{0.210}{0.081}&\resultpair{0.219}{0.074}&\resultpair{0.164}{0.034}\\
\resultgroup{7}{General foundation models}
MGM & \resultpair{0.304}{0.109}&\resultpair{0.272}{0.085}&\resultpair{0.315}{0.127}&\resultpair{0.293}{0.094}&\resultpair{0.271}{0.089}&\resultpair{0.209}{0.054}\\
Waypoint-45M & \resultpair{0.328}{0.121}&\resultpair{0.263}{0.083}&\resultpair{0.331}{0.129}&\resultpair{0.308}{0.106}&\resultpair{0.286}{0.102}&\resultpair{0.239}{0.063}\\
\resultgroup{7}{Gut foundation models}
Pope et al. & \resultpair{0.322}{0.122}&\resultpair{0.286}{0.082}&\resultpair{\underline{0.391}}{\underline{0.183}}&\resultpair{0.259}{0.084}&\resultpair{0.263}{0.094}&\resultpair{\underline{0.258}}{\underline{0.077}}\\
BiomeGPT & \resultpair{0.346}{0.138}&\resultpair{\underline{0.322}}{\underline{0.108}}&\resultpair{0.358}{0.151}&\resultpair{\underline{0.329}}{\underline{0.128}}&\resultpair{\underline{0.322}}{\underline{0.128}}&\resultpair{0.231}{0.058}\\
GUT-FORMer & \resultpair{0.348}{\underline{0.159}}&\resultpair{0.308}{0.091}&\resultpair{0.336}{0.165}&\resultpair{0.303}{0.101}&\resultpair{0.258}{0.088}&\resultpair{0.194}{0.038}\\
MBEmbed & \resultpair{\underline{0.366}}{0.134}&\resultpair{0.311}{0.096}&\resultpair{0.378}{0.144}&\resultpair{0.318}{0.103}&\resultpair{0.301}{0.105}&\resultpair{0.247}{0.069}\\
\specialrule{0.75pt}{2pt}{1pt}
\rowcolor{resultgreen!8}\textcolor{resultgreen}{\textbf{TaxoMFM (ours)}} &
\resultpair{\textbf{0.382}}{\textbf{0.173}}&\resultpair{\textbf{0.337}}{\textbf{0.141}}&\resultpair{\textbf{0.414}}{\textbf{0.202}}&
\resultpair{\textbf{0.347}}{\textbf{0.136}}&\resultpair{\textbf{0.345}}{\textbf{0.142}}&\resultpair{\textbf{0.287}}{\textbf{0.101}}\\
\bottomrule
\end{tabular*}
\end{table}
\endgroup

Table~\ref{tab:classification-results} and Table~\ref{tab:regression-results} reveal a consistent but task-dependent pattern. On public disease tasks, several foundation models already approach high performance, so the gains of TaxoMFM are generally modest. The separation is clearer on gut-health outcomes, especially gastrointestinal sensitivity and gastro-oesophageal reflux disease, where the fine-grained clinical labels are more challenging for the baselines. Regression remains difficult for all methods, as reflected by the moderate Spearman $\rho$ and $R^2$ values, but TaxoMFM improves consistently across host traits and liver ultrasound phenotypes. Overall, the benefit is most visible on clinically detailed and continuous phenotypes rather than on the more saturated disease classification tasks.

\subsection{Ablation study}
\label{sec:ablation-study}

We compare six variants. \textbf{Species-only} removes the genus, family, and order nodes and uses a flat species input. \textbf{Without relation bias} keeps all four taxonomic ranks but removes the relation-dependent attention bias. \textbf{Random taxonomy} preserves the rank sizes and branching pattern but shuffles the parent--child assignments. \textbf{Without hierarchical composition learning} removes the direct supervision of child distributions within each parent. \textbf{Without multiscale reconstruction} removes the reconstruction objective at the four taxonomic ranks. Finally, \textbf{without HPP pretraining} retains the full model and both objectives but excludes HPP samples from pretraining.

\begingroup
\definecolor{ablationgreen}{RGB}{0,104,99}
\definecolor{ablationgroupgray}{RGB}{78,78,78}
\newcommand{\ablationgroup}[2]{%
  \specialrule{0.35pt}{2.2pt}{0pt}
  \rowcolor{black!7}\multicolumn{#1}{@{}l}{\textcolor{ablationgroupgray}{\textbf{#2}}}\\
  \addlinespace[0.5pt]
}
\begin{table}[H]
\caption{Ablation study of TaxoMFM. Classification tasks report AUROC ($\uparrow$), and liver phenotype regression reports Spearman's $\rho$ / $R^2$ ($\uparrow$).}
\label{tab:taxomfm-ablation}
\centering
\fontsize{7.35}{8.2}\selectfont
\setlength{\tabcolsep}{2.7pt}
\renewcommand{\arraystretch}{1.08}

\resizebox{\textwidth}{!}{%
\begin{tabular}{
@{}l
*{3}{c}@{\hspace{5pt}}
*{3}{c}@{\hspace{5pt}}
*{3}{c}@{}
}
\toprule
\multirow{2}{*}{\textbf{Model variant}} &
\multicolumn{3}{c}{\textbf{Disease classification}} &
\multicolumn{3}{c}{\textbf{Gut health classification}} &
\multicolumn{3}{c}{\textbf{Liver phenotype regression}} \\
\cmidrule(lr){2-4}
\cmidrule(lr){5-7}
\cmidrule(l){8-10}

& \shortstack{\textbf{Colorectal}\\\textbf{cancer}}
& \textbf{IBD}
& \shortstack{\textbf{Type 2}\\\textbf{diabetes}}
& \shortstack{\textbf{Current}\\\textbf{abdominal pain}}
& \shortstack{\textbf{Current}\\\textbf{bloating}}
& \shortstack{\textbf{Irritable bowel}\\\textbf{syndrome}}
& \shortstack{\textbf{Attenuation}\\\textbf{coefficient}}
& \shortstack{\textbf{Speed of}\\\textbf{sound}}
& \textbf{Velocity} \\
\midrule

\ablationgroup{10}{Structural ablations}

Species-only (flat)
& 0.914 & 0.943 & 0.909
& 0.736 & 0.758 & 0.723
& 0.313/0.113& 0.320/0.125 & 0.226/0.060 \\

w/o relation bias
& 0.934 & 0.964 & 0.926
& \underline{0.771}& \underline{0.782}& 0.715
& 0.327/0.123& 0.331/0.130& \underline{0.273}/\underline{0.088}\\

Random taxonomy
& 0.906 & 0.933 & 0.880
& 0.686& 0.694& 0.635
& 0.223/0.084& 0.206/0.071& 0.158/0.035 \\

\ablationgroup{10}{Pretraining objective ablations}

w/o hierarchical composition learning
& 0.927 & 0.973 & 0.919
& 0.764& 0.763& 0.719
& \underline{0.331}/\underline{0.129}& \underline{0.339}/\underline{0.134} & 0.271/0.084 \\

w/o multiscale reconstruction
& 0.920 & 0.966 & 0.914
& 0.748 & 0.773 & \underline{0.733}
& 0.320/0.107& 0.332/0.132& 0.252/0.074\\

\ablationgroup{10}{Pretraining data ablation}

w/o HPP pretraining
& \underline{0.946} & \underline{0.980} & \underline{0.928}
& 0.751 & 0.766 & 0.722 
& 0.316/0.105& 0.317/0.109& 0.233/0.068\\

\specialrule{0.75pt}{2pt}{1pt}
\rowcolor{ablationgreen!8}
\textcolor{ablationgreen}{\textbf{TaxoMFM (full)}}
& \textbf{0.953}
& \textbf{0.985}
& \textbf{0.934}
& \textbf{0.776}
& \textbf{0.803}
& \textbf{0.745}
& \textbf{0.347}\,/\,\textbf{0.136}
& \textbf{0.345}\,/\,\textbf{0.142}
& \textbf{0.287}\,/\,\textbf{0.101} \\

\bottomrule
\end{tabular}%
}
\end{table}
\endgroup

The ablation results show that TaxoMFM benefits jointly from taxonomic structure, pretraining objectives, and data coverage. Replacing the true taxonomy with a random hierarchy that preserves rank sizes and branching patterns caused the largest performance drop. The mean AUROC decreased from 0.957 to 0.906 on the three disease tasks and from 0.775 to 0.672 on the three gut-health tasks. The mean Spearman's $\rho$ on the three liver phenotypes decreased from 0.326 to 0.196. The random hierarchy also performed worse than the species-only model, which achieved corresponding group means of 0.922, 0.739, and 0.286. This result shows that the improvement does not come from simply adding higher-rank nodes. It depends on the real taxonomic relationships among microorganisms. Removing the relation bias reduced the mean performance of the three task groups by 0.016, 0.019, and 0.016, respectively, showing an additional benefit from explicitly modeling taxonomic relations. The two pretraining objectives also provided complementary supervision. Removing multiscale reconstruction reduced the mean performance by 0.024, 0.023, and 0.025 across the three task groups, while removing hierarchical composition learning resulted in decreases of 0.018, 0.026, and 0.013. Finally, excluding HPP from pretraining reduced the mean AUROC on public disease tasks by only 0.006, but reduced the mean AUROC on gut-health tasks by 0.028 and the mean Spearman's $\rho$ on liver phenotypes by 0.038. These results show that HPP primarily expands the model's pretraining coverage of gut health and continuous clinical phenotypes.

\subsection{Zero-shot cross-cohort transfer}

We evaluate target-cohort zero-shot transfer on three external disease cohorts: colorectal cancer from Yang et al. (2021), type 2 diabetes from MetaCardis (2020), and IBD from HMP/IBDMDB. For each task, the target cohort is excluded from both pretraining and downstream adaptation. Models are trained using only labeled source cohorts and are then evaluated directly on the unseen target cohort using AUROC. As shown in Table~\ref{tab:strict-zero-shot}, TaxoMFM achieves the best performance on all three tasks, with AUROCs of 0.838, 0.736, and 0.722, respectively. Compared with the strongest baseline for each task, TaxoMFM improves AUROC by 0.075 on colorectal cancer, 0.039 on type 2 diabetes, and 0.029 on IBD. These results show that taxonomy-aware pretraining improves cross-cohort transfer across cancer, metabolic disease, and inflammatory disease.

\subsection{Analysis of taxonomy-aware pretraining}

The previous experiments show that taxonomic structure improves downstream performance. We next examine what the model learns from this structure through three targeted analyses.
\vspace{-3pt}
\paragraph{Distinguishing changes across taxonomic scales.}
We first test whether TaxoMFM can recover community changes at different taxonomic scales from incomplete observations. We use only samples excluded from model pretraining. For each sample, we select a genus containing multiple detected species and construct three controlled changes: changing a single species, changing the total abundance of the genus, or redistributing abundance among its species while keeping the genus total fixed. For each original--perturbed pair, we mask the same species within the selected genus and reconstruct both complete communities. We then compare the reconstructed difference with the known perturbation. Parent error measures the recovery of genus-level abundance changes, conditional error measures the recovery of within-genus compositional changes, and localization error measures whether the change is assigned to the correct genus. Combined error, our primary metric, is the sum of the parent and conditional errors.

\vspace{-8pt}
\begingroup
\definecolor{analysisgreen}{RGB}{0,104,99}
\begin{figure}[H]
\centering
\begin{minipage}[c]{0.465\textwidth}
\captionof{table}{Zero-shot performance using \mbox{AUROC} ($\uparrow$).}
\label{tab:strict-zero-shot}
\centering
\fontsize{7.3}{8.4}\selectfont
\setlength{\tabcolsep}{2.4pt}
\renewcommand{\arraystretch}{1.05}
\begin{tabular}{@{}>{\raggedright\arraybackslash}p{0.305\linewidth}*{3}{>{\centering\arraybackslash}p{0.20\linewidth}}@{}}
\toprule
\textbf{Model} &
\shortstack{\textbf{CRC}\\\scalebox{0.95}{\textbf{Yang et al.}}\\\textbf{(2021)}\\$n=194$} &
\shortstack{\textbf{T2D}\\\scalebox{0.90}{\textbf{MetaCardis}}\\\textbf{(2020)}\\$n=783$} &
\shortstack{\textbf{IBD}\\\textbf{HMP/}\\\textbf{IBDMDB}\\$n=130$} \\
\midrule
Random Forest & 0.619 & 0.634  & 0.631  \\
MIOSTONE & 0.643  & 0.562  &  \underline{0.693} \\
PhyloGCNE & 0.610 & 0.602  & 0.567  \\
MGM & 0.744  & 0.670  & 0.664  \\
BiomeGPT &  \underline{0.763} & 0.683  & 0.679  \\
Species-only &  0.759 & \underline{0.697}  & 0.688  \\
\specialrule{0.7pt}{2pt}{1pt}
\rowcolor{analysisgreen!8}
\textcolor{analysisgreen}{\scalebox{0.90}{\textbf{TaxoMFM (ours)}}} &
 \textbf{0.838}   & \textbf{0.736} & \textbf{0.722}   \\
\bottomrule
\end{tabular}
\end{minipage}
\hfill
\begin{minipage}[c]{0.515\textwidth}
\captionof{table}{Multiscale compositional recovery under controlled perturbations ($\downarrow$).}
\label{tab:multiscale-recovery}
\centering
\fontsize{8.0}{9.0}\selectfont
\setlength{\tabcolsep}{1.4pt}
\renewcommand{\arraystretch}{1.08}
\resizebox{\linewidth}{!}{%
\begin{tabular}{@{}lcccc@{}}
\toprule
\textbf{Model variant} &
\textbf{Parent} &
\textbf{Conditional} &
\textbf{Localization} &
\textbf{Combined} \\
\midrule
Species-only (flat) & 0.736 & \underline{0.994} & \underline{0.489} & \underline{1.730} \\
Random taxonomy & 0.823 & 1.414 & 0.621 & 2.237 \\
w/o hierarchical & \multirow{2}{*}{\underline{0.722}} & \multirow{2}{*}{1.228} & \multirow{2}{*}{0.525} & \multirow{2}{*}{1.950} \\[-1.5pt]
composition learning & & & & \\
\specialrule{0.7pt}{2pt}{1pt}
\rowcolor{analysisgreen!8}
\textcolor{analysisgreen}{\textbf{TaxoMFM (full)}} &
\textcolor{analysisgreen}{\textbf{0.642}} &
\textcolor{analysisgreen}{\textbf{0.956}} &
\textcolor{analysisgreen}{\textbf{0.431}} &
\textcolor{analysisgreen}{\textbf{1.598}} \\
\bottomrule
\end{tabular}%
}
\end{minipage}
\end{figure}
\endgroup
\vspace{-4pt}

As shown in Table~\ref{tab:multiscale-recovery}, TaxoMFM achieves the lowest error on all four metrics. Its combined error is 1.598, representing reductions of 7.6\%, 28.6\%, and 18.1\% relative to species-only, random taxonomy, and the model without hierarchical composition learning, respectively. The lower parent and conditional errors indicate that TaxoMFM better recovers both lineage-level abundance changes and within-lineage compositional shifts. Performance declines when the taxonomy is randomized or hierarchical composition learning is removed, indicating that both components contribute to multiscale compositional recovery.
\vspace{-3pt}
\paragraph{Conditional recovery of long-tail species.}
We evaluated conditional abundance recovery on independent external samples. Species were divided into four groups according to their prevalence in the pretraining data, from Q1 (rarest) to Q4 (most prevalent). For each instance, we masked the abundances of approximately 20\% of detected species, including the target and at least one additional species. We retained species identities and constructed model inputs from the renormalized visible composition. The prediction target was the target species' abundance divided by the total visible abundance. We measured recovery using species-averaged absolute $\log_2$-ratio error, with lower values indicating better performance.

\begingroup
\definecolor{analysisgreen}{RGB}{0,104,99}
\begin{figure}[!t]
\centering
\begin{minipage}[t]{0.46\textwidth}
\vspace{0pt}
\begin{minipage}[t][49pt][t]{\linewidth}
\captionof{table}{Long-tail recovery from Q1 (rarest) to Q4 (commonest).}
\label{tab:longtail-recovery}
\end{minipage}
\par\centering
\fontencoding{OT1}\fontfamily{ptm}\selectfont
\fontsize{7.3}{9.2}\selectfont
\setlength{\tabcolsep}{2.4pt}
\renewcommand{\arraystretch}{1.34}
\begin{tabular}{@{}lrrrr@{}}
\toprule
\textbf{Method} & \textbf{Q1} & \textbf{Q2} & \textbf{Q3} & \textbf{Q4} \\
\midrule
Species identity prior & \underline{5.153} & \underline{7.342} & 7.563 & \underline{5.284} \\
Genus proportion prior & 6.337 & 8.624 & \underline{7.118} & 6.074 \\
\midrule
Species-only & 7.012 & 9.483 & 8.661 & 6.653 \\
Random taxonomy & 23.126 & 19.847 & 21.203 & 13.014 \\
w/o hierarchical loss & 13.924 & 11.718 & 12.306 & 9.347 \\
BiomeGPT & 5.881 & 7.832 & 7.384 & 6.422 \\
\specialrule{0.75pt}{2pt}{1pt}
\rowcolor{analysisgreen!8}
\textcolor{analysisgreen}{\textbf{TaxoMFM}} & \textbf{4.139} & \textbf{6.479} & \textbf{6.934} & \textbf{4.866} \\
\bottomrule
\end{tabular}
\end{minipage}
\hfill
\begin{minipage}[t]{0.52\textwidth}
\vspace{0pt}
\par\centering
\begin{minipage}[t][49pt][t]{\linewidth}
\captionof{figure}{Recovery trends across prevalence groups.}
\label{fig:longtail-recovery}
\end{minipage}
\includegraphics[width=\linewidth]{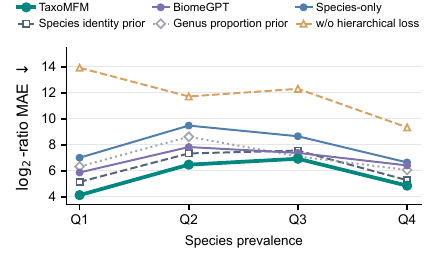}
\end{minipage}
\end{figure}
\endgroup

Table~\ref{tab:longtail-recovery} and Fig.~\ref{fig:longtail-recovery} show TaxoMFM has the lowest error across all four prevalence groups. Its errors are 29.6\%, 17.3\%, 6.1\%, and 24.2\% lower than the BiomeGPT values in Q1–Q4, respectively, with the largest differences in Q1 and Q4. In Q1, TaxoMFM’s error is 4.139, compared with 5.153 for the species identity prior and 6.337 for the genus proportion prior. Baseline rankings vary across groups: the species identity prior performs best among baselines in Q1, Q2, and Q4, whereas the genus proportion prior performs best in Q3.
\vspace{-7pt}
\paragraph{Maintaining representations across taxonomic vocabularies.}
Different taxonomic database versions can rename, split, or omit the same species or SGB. We therefore constructed paired MetaPhlAn 4 current and MetaPhlAn 3 legacy views from profiles in the pretraining dataset. The legacy view retained identical species names and 33 curated MetaPhlAn 3-to-4 mappings, whereas TaxoMFM routed unmatched abundance to genus, family, or order nodes shared across the vocabularies. Because each pair came from the same sample, only the vocabulary changed.

\vspace{-6pt}
\begingroup
\definecolor{analysisgreen}{RGB}{0,104,99}
\begin{figure}[H]
\centering
\begin{minipage}[t]{0.485\textwidth}
\vspace{0pt}
\captionof{table}{Cross-vocabulary consistency under a MetaPhlAn 3-to-4 shift.}
\label{tab:cross-vocabulary}
\centering
\fontsize{7.6}{8.5}\selectfont
\setlength{\tabcolsep}{2.6pt}
\renewcommand{\arraystretch}{1.08}
\resizebox{\linewidth}{!}{%
\begin{tabular}{@{}lcccc@{}}
\toprule
\textbf{Model} &
\shortstack{\textbf{Legacy abundance}\\\textbf{retained (\%) $\uparrow$}} &
\shortstack{\textbf{Cosine}\\\textbf{separation $\uparrow$}} &
\shortstack{\textbf{Bidirectional}\\\textbf{R@1 (\%) $\uparrow$}} &
\shortstack{\textbf{Bidirectional}\\\textbf{R@5 (\%) $\uparrow$}} \\
\midrule
\rowcolor{black!7}
\multicolumn{5}{@{}l}{\textit{Structural controls}} \\
TaxoMFM leaf-only & 45.79 & 0.309 & 67.69 & 84.94 \\
Species-only (flat) & 45.79 & \underline{0.370} & 82.42 & 84.52 \\
Random taxonomy & \textbf{81.44} & 0.045 & 1.23 & 3.51 \\
\midrule
\rowcolor{black!7}
\multicolumn{5}{@{}l}{\textit{External baselines}} \\
BiomeGPT & 45.79 & 0.078 & 83.11 & 83.99 \\
Waypoint-45M & \underline{64.30} & 0.116 & 81.06 & 83.54 \\
GUT-FORMer & 41.05 & 0.050 & 79.53 & 88.84 \\
MBEmbed & 33.14 & 0.290 & 75.32 & 89.18 \\
DeepMicro & 41.05 & 0.104 & \underline{85.73} & \underline{90.79} \\
\specialrule{0.75pt}{2pt}{1pt}
\rowcolor{analysisgreen!8}
\textcolor{analysisgreen}{\textbf{TaxoMFM (full)}} &
\textcolor{analysisgreen}{\textbf{81.44}} &
\textcolor{analysisgreen}{\textbf{0.379}} &
\textcolor{analysisgreen}{\textbf{86.64}} &
\textcolor{analysisgreen}{\textbf{96.06}} \\
\bottomrule
\end{tabular}%
}
\end{minipage}
\hfill
\begin{minipage}[t]{0.485\textwidth}
\vspace{0pt}
\centering
\begin{minipage}[t][49pt][t]{\linewidth}
\captionof{figure}{Cross-vocabulary transfer performance across models, with darker cells indicating higher within-metric rank.}
\label{fig:cross-vocabulary-heatmap}
\end{minipage}
\includegraphics[width=\linewidth]{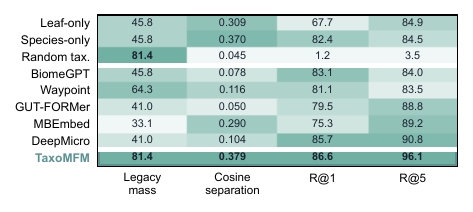}
\end{minipage}
\end{figure}
\endgroup
\vspace{-6pt}

We report retained legacy abundance (the mean fraction of original
abundance preserved through species matching or ancestor routing),
cosine separation, and bidirectional Recall@1 and Recall@5.
Detailed metric definitions are provided in
Appendix~\ref{sec:appendix-vocabulary-settings}.
As shown in Table~\ref{tab:cross-vocabulary} and Fig.~\ref{fig:cross-vocabulary-heatmap}, TaxoMFM retained 81.44\% of the abundance and achieved a cosine separation of 0.379, a Recall@1 of 86.64\%, and a Recall@5 of 96.06\%. Disabling ancestor routing reduced retained abundance to 45.79\% and Recall@1 to 67.69\%. A random taxonomy retained the same abundance but achieved only 1.23\% Recall@1. Shared ancestors therefore preserve information across vocabularies, while the correct taxonomic structure maintains sample-specific representations.

\section{Conclusion}
We introduced TaxoMFM, a microbiome foundation model that incorporates taxonomic hierarchies and relationships across species, genus, family, and order. Experiments on TaxoGut-HPP show improvements in disease classification, gut health assessment, continuous phenotype prediction, and cross-cohort zero-shot transfer. TaxoMFM also recovers community changes across taxonomic scales and masked species abundances while adapting to vocabulary shifts through shared taxonomic ancestors. Together, these results establish a framework for learning biologically informed microbiome representations that transfer across tasks, cohorts, and taxonomic vocabularies.
\fi

\clearpage

\subsection*{AI use statement}
We used large language models (LLMs) to assist with dataset
collection, manuscript polishing, and code review.
The authors take responsibility for the final content of this work.

\bibliography{iclr2027_conference}

\begin{thebibliography}{21}
\providecommand{\natexlab}[1]{#1}
\providecommand{\url}[1]{\texttt{#1}}
\expandafter\ifx\csname urlstyle\endcsname\relax
  \providecommand{\doi}[1]{doi: #1}\else
  \providecommand{\doi}{doi: \begingroup \urlstyle{rm}\Url}\fi

\bibitem[Aitchison(1986)]{aitchison1986statistical}
John Aitchison.
\newblock \emph{The Statistical Analysis of Compositional Data}.
\newblock Chapman and Hall, 1986.

\bibitem[Bommasani et~al.(2021)Bommasani, Hudson, Adeli, Altman, Arora, von
  Arx, Bernstein, Bohg, Bosselut, Brunskill, et~al.]{bommasani2021foundation}
Rishi Bommasani, Drew~A. Hudson, Ehsan Adeli, Russ Altman, Simran Arora, Sydney
  von Arx, Michael~S. Bernstein, Jeannette Bohg, Antoine Bosselut, Emma
  Brunskill, et~al.
\newblock On the opportunities and risks of foundation models.
\newblock \emph{arXiv preprint arXiv:2108.07258}, 2021.
\newblock \doi{10.48550/arXiv.2108.07258}.

\bibitem[Chang et~al.(2024)Chang, Gupta, Hur, Cobo-L{\'o}pez, Cunningham, Han,
  Lee, Kronzer, Teigen, Karnatovskaia, et~al.]{chang2024gut}
Daniel Chang, Vinod~K Gupta, Benjamin Hur, Sergio Cobo-L{\'o}pez, Kevin~Y
  Cunningham, Nam~Soo Han, Insuk Lee, Vanessa~L Kronzer, Levi~M Teigen,
  Lioudmila~V Karnatovskaia, et~al.
\newblock Gut microbiome wellness index 2 enhances health status prediction
  from gut microbiome taxonomic profiles.
\newblock \emph{Nature communications}, 15\penalty0 (1):\penalty0 7447, 2024.

\bibitem[Dong et~al.(2026)Dong, Wang, Chen, Xu, and Xu]{dong2026phylogcne}
Biao Dong, Bin Wang, Jiongjin Chen, Xiaomin Xu, and Zhenjiang~Zech Xu.
\newblock Adaptive graph learning of microbial phylogeny enables accurate and
  interpretable microbiome-based host phenotype prediction.
\newblock \emph{Applied and Environmental Microbiology}, 2026.
\newblock \doi{10.1128/aem.00788-26}.

\bibitem[Fioravanti et~al.(2018)Fioravanti, Giarratano, Maggio, Agostinelli,
  Chierici, Jurman, and Furlanello]{fioravanti2018phcnn}
Diego Fioravanti, Ylenia Giarratano, Valerio Maggio, Claudio Agostinelli, Marco
  Chierici, Giuseppe Jurman, and Cesare Furlanello.
\newblock Phylogenetic convolutional neural networks in metagenomics.
\newblock \emph{BMC Bioinformatics}, 19:\penalty0 49, 2018.
\newblock \doi{10.1186/s12859-018-2033-5}.

\bibitem[Forslund et~al.(2021)Forslund, Chakaroun, Zimmermann-Kogadeeva,
  Markó, Aron-Wisnewsky, Nielsen, Moitinho-Silva, Schmidt, Falony,
  Vieira-Silva, Adriouch, Alves, Assmann, Bastard, Birkner, Caesar, Chilloux,
  Coelho, Fezeu, Galleron, Helft, Isnard, Ji, Kuhn, Le~Chatelier, Myridakis,
  Olsson, Pons, Prifti, Quinquis, Roume, Salem, Sokolovska, Tremaroli,
  Valles-Colomer, Lewinter, Søndertoft, Pedersen, Hansen, {MetaCardis
  Consortium*}, Gøtze, Køber, Vestergaard, Hansen, Zucker, Hercberg, Oppert,
  Letunic, Nielsen, Bäckhed, Ehrlich, Dumas, Raes, Pedersen, Clément,
  Stumvoll, and Bork]{forslund2021metacardis}
Sofia~K Forslund, Rima Chakaroun, Maria Zimmermann-Kogadeeva, Lajos Markó,
  Judith Aron-Wisnewsky, Trine Nielsen, Lucas Moitinho-Silva, Thomas S~B
  Schmidt, Gwen Falony, Sara Vieira-Silva, Solia Adriouch, Renato~J Alves,
  Karen Assmann, Jean-Philippe Bastard, Till Birkner, Robert Caesar, Julien
  Chilloux, Luis~Pedro Coelho, Leopold Fezeu, Nathalie Galleron, Gerard Helft,
  Richard Isnard, Boyang Ji, Michael Kuhn, Emmanuelle Le~Chatelier, Antonis
  Myridakis, Lisa Olsson, Nicolas Pons, Edi Prifti, Benoit Quinquis, Hugo
  Roume, Joe-Elie Salem, Nataliya Sokolovska, Valentina Tremaroli, Mireia
  Valles-Colomer, Christian Lewinter, Nadja~B Søndertoft, Helle~Krogh
  Pedersen, Tue~H Hansen, {MetaCardis Consortium*}, Jens~Peter Gøtze, Lars
  Køber, Henrik Vestergaard, Torben Hansen, Jean-Daniel Zucker, Serge
  Hercberg, Jean-Michel Oppert, Ivica Letunic, Jens Nielsen, Fredrik Bäckhed,
  S~Dusko Ehrlich, Marc-Emmanuel Dumas, Jeroen Raes, Oluf Pedersen, Karine
  Clément, Michael Stumvoll, and Peer Bork.
\newblock {Combinatorial, additive and dose-dependent drug-microbiome
  associations}.
\newblock \emph{Nature}, 600\penalty0 (7889):\penalty0 500--505, 2021.
\newblock \doi{10.1038/s41586-021-04177-9}.

\bibitem[Jiang et~al.(2025)Jiang, Aton, Zhu, and Lu]{jiang2025miostone}
Yifan Jiang, Matthew Aton, Qiyun Zhu, and Yang~Young Lu.
\newblock Modeling microbiome-trait associations with taxonomy-adaptive neural
  networks.
\newblock \emph{Microbiome}, 13\penalty0 (1), 2025.
\newblock \doi{10.1186/s40168-025-02080-3}.

\bibitem[Karwowska et~al.(2026)Karwowska, Mo{\.{z}}ejko, Nowak, Romanchenko,
  Szczurek, and Kosciolek]{karwowska2026gutformer}
Zuzanna Karwowska, Marcin Mo{\.{z}}ejko, Wojciech Nowak, Anastasia Romanchenko,
  Ewa Szczurek, and Tomasz Kosciolek.
\newblock A universal taxonomic and functional human gut microbiome model for
  disease classification and phenotype discovery.
\newblock \emph{bioRxiv}, 2026.
\newblock \doi{10.64898/2026.04.30.721924}.

\bibitem[Lloyd-Price et~al.(2019)Lloyd-Price, Arze, Ananthakrishnan, Schirmer,
  Avila-Pacheco, Poon, Andrews, Ajami, Bonham, Brislawn, Casero, Courtney,
  Gonzalez, Graeber, Hall, Lake, Landers, Mallick, Plichta, Prasad, Rahnavard,
  Sauk, Shungin, Vázquez-Baeza, White, {IBDMDB Investigators}, Braun, Denson,
  Jansson, Knight, Kugathasan, McGovern, Petrosino, Stappenbeck, Winter, Clish,
  Franzosa, Vlamakis, Xavier, and Huttenhower]{lloydprice2019ibdmdb}
Jason Lloyd-Price, Cesar Arze, Ashwin~N Ananthakrishnan, Melanie Schirmer,
  Julian Avila-Pacheco, Tiffany~W Poon, Elizabeth Andrews, Nadim~J Ajami,
  Kevin~S Bonham, Colin~J Brislawn, David Casero, Holly Courtney, Antonio
  Gonzalez, Thomas~G Graeber, A~Brantley Hall, Kathleen Lake, Carol~J Landers,
  Himel Mallick, Damian~R Plichta, Mahadev Prasad, Gholamali Rahnavard, Jenny
  Sauk, Dmitry Shungin, Yoshiki Vázquez-Baeza, Richard~A White, {IBDMDB
  Investigators}, Jonathan Braun, Lee~A Denson, Janet~K Jansson, Rob Knight,
  Subra Kugathasan, Dermot P~B McGovern, Joseph~F Petrosino, Thaddeus~S
  Stappenbeck, Harland~S Winter, Clary~B Clish, Eric~A Franzosa, Hera Vlamakis,
  Ramnik~J Xavier, and Curtis Huttenhower.
\newblock {Multi-omics of the gut microbial ecosystem in inflammatory bowel
  diseases}.
\newblock \emph{Nature}, 569\penalty0 (7758):\penalty0 655--662, 2019.
\newblock \doi{10.1038/s41586-019-1237-9}.

\bibitem[Medearis et~al.(2026)Medearis, Zhu, and
  Zomorrodi]{medearis2026biomegpt}
Nicholas~A. Medearis, Siyao Zhu, and Ali~R. Zomorrodi.
\newblock {BiomeGPT}: A foundation model for the human gut microbiome.
\newblock \emph{bioRxiv}, 2026.
\newblock \doi{10.64898/2026.01.05.697599}.

\bibitem[Oh \& Zhang(2020)Oh and Zhang]{oh2020deepmicro}
Min Oh and Liqing Zhang.
\newblock {DeepMicro}: deep representation learning for disease prediction
  based on microbiome data.
\newblock \emph{Scientific Reports}, 10:\penalty0 6026, 2020.
\newblock \doi{10.1038/s41598-020-63159-5}.

\bibitem[Pasolli et~al.(2017)Pasolli, Schiffer, Manghi, Renson, Obenchain,
  Truong, Beghini, Malik, Ramos, Dowd, et~al.]{pasolli2017accessible}
Edoardo Pasolli, Lucas Schiffer, Paolo Manghi, Audrey Renson, Valerie
  Obenchain, Duy~Tin Truong, Francesco Beghini, Faizan Malik, Marcel Ramos,
  Jennifer~B Dowd, et~al.
\newblock Accessible, curated metagenomic data through experimenthub.
\newblock \emph{Nature methods}, 14\penalty0 (11):\penalty0 1023--1024, 2017.

\bibitem[Pope et~al.(2025)Pope, Varma, Tataru, David, and
  Fern]{pope2025microbiomes}
Quintin Pope, Rohan Varma, Christine Tataru, Maude~M. David, and Xiaoli Fern.
\newblock Learning a deep language model for microbiomes: The power of large
  scale unlabeled microbiome data.
\newblock \emph{PLOS Computational Biology}, 21\penalty0 (5):\penalty0
  e1011353, 2025.
\newblock \doi{10.1371/journal.pcbi.1011353}.

\bibitem[Reiman et~al.(2020)Reiman, Metwally, Sun, and
  Dai]{reiman2020popphycnn}
Derek Reiman, Ahmed~A. Metwally, Jun Sun, and Yang Dai.
\newblock {PopPhy-CNN}: a phylogenetic tree embedded architecture for
  convolutional neural networks to predict host phenotype from metagenomic
  data.
\newblock \emph{IEEE Journal of Biomedical and Health Informatics}, 24\penalty0
  (10):\penalty0 2993--3001, 2020.
\newblock \doi{10.1109/JBHI.2020.2993761}.

\bibitem[Sharma et~al.(2020)Sharma, Paterson, and Xu]{sharma2020taxonn}
Divya Sharma, Andrew~D. Paterson, and Wei Xu.
\newblock {TaxoNN}: ensemble of neural networks on stratified microbiome data
  for disease prediction.
\newblock \emph{Bioinformatics}, 36\penalty0 (17):\penalty0 4544--4550, 2020.
\newblock \doi{10.1093/bioinformatics/btaa542}.

\bibitem[Silverman et~al.(2017)Silverman, Washburne, Mukherjee, and
  David]{silverman2017philr}
Justin~D. Silverman, Alex~D. Washburne, Sayan Mukherjee, and Lawrence~A. David.
\newblock A phylogenetic transform enhances analysis of compositional
  microbiota data.
\newblock \emph{eLife}, 6:\penalty0 e21887, 2017.
\newblock \doi{10.7554/eLife.21887}.

\bibitem[Treloar et~al.(2026)Treloar, Ur-Rehman, and Yang]{treloar2026waypoint}
Neythen~J. Treloar, Saif Ur-Rehman, and Jenny Yang.
\newblock Learning the language of the microbiome with transformers.
\newblock \emph{bioRxiv}, 2026.
\newblock \doi{10.64898/2026.05.02.722381}.

\bibitem[Wang et~al.(2024)Wang, Shen, Fang, and Su]{wang2024deepphylo}
Bin Wang, Yulong Shen, Jingyan Fang, and Xiaoquan Su.
\newblock {DeepPhylo}: phylogeny-aware microbial embeddings enhanced predictive
  accuracy in human microbiome data analysis.
\newblock \emph{Advanced Science}, 11\penalty0 (45):\penalty0 2404277, 2024.
\newblock \doi{10.1002/advs.202404277}.

\bibitem[Yang et~al.(2021)Yang, Du, Shi, Kong, Liu, Liu, Li, and
  Ma]{yang2021youngcrc}
Yongzhi Yang, Lutao Du, Debing Shi, Cheng Kong, Jianqiang Liu, Guang Liu,
  Xinxiang Li, and Yanlei Ma.
\newblock {Dysbiosis of human gut microbiome in young-onset colorectal cancer}.
\newblock \emph{Nature Communications}, 12\penalty0 (1):\penalty0 6757, 2021.
\newblock \doi{10.1038/s41467-021-27112-y}.

\bibitem[Zahavi et~al.(2025)Zahavi, Levine, and Segal]{zahavi2025mbembed}
Liron Zahavi, Zachary Levine, and Eran Segal.
\newblock Self-supervised representation learning for microbiome improves
  downstream prediction in data-limited settings and cross-cohort
  generalizability.
\newblock In \emph{ICML Workshop on Multi-modal Foundation Models and Large
  Language Models for Life Sciences}, 2025.

\bibitem[Zhang et~al.(2026)Zhang, Zhang, Kang, Xiong, Yang, and
  Ning]{zhang2026mgm}
Haohong Zhang, Yuli Zhang, Zixin Kang, Jiayun Xiong, Ronghua Yang, and Kang
  Ning.
\newblock {MGM} as a large-scale pretrained foundation model for microbiome
  analyses in diverse contexts.
\newblock \emph{Advanced Science}, pp.\  e13333, 2026.
\newblock \doi{10.1002/advs.202513333}.

\end{thebibliography}
\bibliographystyle{iclr2027_conference}

\clearpage
\appendix
\setcounter{table}{0}
\renewcommand{\thetable}{A\arabic{table}}
\renewcommand{\theHtable}{appendix.A\arabic{table}}
\section{Additional downstream results}
\label{sec:additional_downstream_results}
\def\experimentappendix{1}

\raggedbottom
\section{Detailed experimental settings}
\label{sec:appendix-experimental-settings}

Cross-cohort evaluation tests whether disease-related information learned from source cohorts transfers to external populations. Tables~\ref{tab:multiscale-recovery}--\ref{tab:cross-vocabulary} further examine the information retained in the representations through community perturbations, conditional recovery of rare species, and vocabulary changes. These experiments assess whether models distinguish compositional changes at different taxonomic scales, recover masked abundances from the visible community, and preserve sample distinctions across vocabularies. We describe the data splits, input construction, evaluation procedures, and metrics below.

\subsection{Zero-shot cross-cohort transfer (Table~\ref{tab:strict-zero-shot})}
\label{sec:appendix-cross-cohort}

We evaluate cross-cohort transfer for colorectal cancer (CRC), type 2 diabetes (T2D), and inflammatory bowel disease (IBD). Each task comprises labeled source cohorts and one external target test set, to which the trained model is applied directly. A separate pretraining split excludes all 1,107 target test samples from pretraining and downstream training. Table~\ref{tab:appendix-transfer-splits} summarizes the task-specific cohort splits.

\begin{table}[H]
\centering
\caption{Source and target cohort splits for cross-cohort transfer. Positive and negative counts refer to the target test sets.}
\label{tab:appendix-transfer-splits}
\small
\setlength{\tabcolsep}{4pt}
\renewcommand{\arraystretch}{1.12}
\begin{tabular*}{\textwidth}{@{\extracolsep{\fill}}lr>{\raggedright\arraybackslash}p{0.42\textwidth}rr@{}}
\toprule
\textbf{Task} & \shortstack{\textbf{Source}\\\textbf{samples}} & \textbf{Target study} & \shortstack{\textbf{Target}\\\textbf{samples}} & \shortstack{\textbf{Positive/}\\\textbf{negative}}\\
\midrule
CRC & 1,517 & \citet{yang2021youngcrc} & 194 & 97/97\\
T2D & 465 & MetaCardis~\citep{forslund2021metacardis} & 783 & 549/234\\
IBD & 318 & HMP/IBDMDB~\citep{lloydprice2019ibdmdb} & 130 & 103/27\\
\bottomrule
\end{tabular*}
\end{table}

The public profiles were obtained from curatedMetagenomicData~\citep{pasolli2017accessible} and GMWI2~\citep{chang2024gut}. Table~\ref{tab:appendix-transfer-provenance} lists the source and target datasets, their resource-specific study identifiers, and the PubMed identifiers of the corresponding original studies. Splits are organized by participant, with one sample per participant in each task.

\begin{table}[H]
\centering
\caption{Resource study IDs and PubMed identifiers for cross-cohort transfer.}
\label{tab:appendix-transfer-provenance}
\small
\setlength{\tabcolsep}{4pt}
\renewcommand{\arraystretch}{1.08}
\begin{tabular*}{\textwidth}{@{\extracolsep{\fill}}llllr@{}}
\toprule
\textbf{Task} & \textbf{Role} & \textbf{Resource} & \textbf{Resource study ID} & \textbf{PMID}\\
\midrule
CRC & Source & cMD & GuptaA\_2019 & 31719139\\
CRC & Source & cMD & HanniganGD\_2017 & 30459201\\
CRC & Source & cMD & ThomasAM\_2018b & 30936548\\
CRC & Source & cMD & VogtmannE\_2016 & 27171425\\
CRC & Source & cMD & WirbelJ\_2018 & 30936547\\
CRC & Source & cMD & YachidaS\_2019 & 31171880\\
CRC & Source & GMWI2 & Thomas\_2019 & 30936548\\
CRC & Source & GMWI2 & Yang\_2020 & 31971861\\
CRC & Source & GMWI2 & Yu\_2015 & 26408641\\
CRC & Source & GMWI2 & Zeller\_2014 & 25432777\\
CRC & Target & GMWI2 & Yang\_2021 & 34799562\\
\midrule
T2D & Source & cMD & KarlssonFH\_2013 & 23719380\\
T2D & Source & GMWI2 & Qin\_2012 & 23023125\\
T2D & Target & cMD & MetaCardis\_2020\_a & 34880489\\
\midrule
IBD & Source & cMD & NielsenHB\_2014 & 24997787\\
IBD & Target & cMD & HMP\_2019\_ibdmdb & 31142855\\
\bottomrule
\end{tabular*}
\end{table}

\subsection{Recovering compositional changes across taxonomic scales (Table~\ref{tab:multiscale-recovery})}
\label{sec:appendix-multiscale-settings}

We construct three magnitude-matched compositional perturbations in samples outside pretraining to assess recovery of changes in lineage abundance and within-lineage species proportions. Let $x$ denote species relative abundances and $\mathcal C_g$ the species in genus $g$, with total abundance $P_g=\sum_{s\in\mathcal C_g}x_s$. We set the transferred abundance to $\delta=\alpha P_g$, with $\alpha=0.10$. The three operations increase a single species by $\delta$, proportionally scale all species in the genus to increase its total by $\delta$, or transfer $\delta$ between two congeneric species. The first two remove an equal amount proportionally from the community outside the genus; the third preserves the genus total. In each original--perturbed pair, we mask the same detected species other than the donor and recipient, retaining its identity and setting its input abundance to zero. We independently renormalize the visible species abundances in each composition to sum to one. All higher-rank input abundances are then recomputed from the renormalized visible composition. The model reconstructs both complete compositions, and their difference estimates the true perturbation. Comparisons include full TaxoMFM, species-only, random taxonomy, and the variant without hierarchical composition learning.

Let $\Delta$ denote the perturbed value minus the original value, with hatted quantities calculated from the reconstructed composition $\hat x$. Writing the within-genus proportion as $q_{s\mid g}=x_s/P_g$, Parent error and Conditional error quantify recovery of changes in genus totals and within-genus proportions, respectively. They are normalized by $\delta$ and $\alpha$:
\begin{equation}
\begin{aligned}
E_{\mathrm{parent}}&=\frac{|\Delta\hat P_g-\Delta P_g|}{\delta},\\
E_{\mathrm{conditional}}&=\frac{1}{\alpha}\sum_{s\in\mathcal C_g}|\Delta\hat q_{s\mid g}-\Delta q_{s\mid g}|.
\end{aligned}
\label{eq:appendix-parent-conditional}
\end{equation}
Localization error compares the fractions of predicted and true absolute changes that occur within the target genus. Denoting this fraction for a change vector $v$ by $L_g(v)$ gives
\begin{equation}
L_g(v)=\frac{\sum_{s\in\mathcal C_g}|v_s|}{\sum_s|v_s|},
\qquad E_{\mathrm{localization}}=|L_g(\Delta\hat x)-L_g(\Delta x)|.
\label{eq:appendix-localization}
\end{equation}
Combined error, $E_{\mathrm{combined}}=E_{\mathrm{parent}}+E_{\mathrm{conditional}}$, aggregates errors in recovering genus totals and within-genus proportions. All metrics are averaged across samples and the three perturbations; lower values indicate better recovery.

\subsection{Conditional abundance recovery of rare species (Table~\ref{tab:longtail-recovery})}
\label{sec:appendix-conditional-settings}

We divide eligible species detected in the external evaluation samples into four approximately equal-sized groups according to their prevalence in the pretraining data. Groups Q1--Q4 range from the rarest to the most common species. In each external sample, a detected target species is masked together with other randomly selected species. Each mask covers approximately 20\% of detected species and retains at least two visible species. All methods use the same targets and masks. Species identities are retained, masked abundances are set to zero, and visible abundances are renormalized; higher-rank inputs are aggregated from visible species. Let $x$ denote the original abundances, $V$ the visible species, and $t$ the target species. The evaluation target is the log abundance ratio of the target species to the visible community total. The model computes the corresponding prediction from its reconstructed composition $\hat p$:
\begin{equation}
\begin{aligned}
\tilde x_s&=\begin{cases}x_s/\sum_{j\in V}x_j,&s\in V,\\0,&s\text{ is masked},\end{cases}\\[3pt]
y_t&=\log_2\frac{x_t}{\sum_{j\in V}x_j},
\qquad\hat y_t=\log_2\frac{\hat p_t}{\sum_{j\in V}\hat p_j}.
\end{aligned}
\label{eq:appendix-conditional-target}
\end{equation}

Table~\ref{tab:longtail-recovery} reports the absolute log-ratio error $e_t=|\hat y_t-y_t|$. We first average repeated masks and repeated samples within participants, then average across participants for each species. Species are weighted equally within each prevalence group; lower errors indicate more accurate recovery. Two priors estimated from the training data provide reference predictions. The Species identity prior uses each species' median conditional log abundance ratio. The Genus proportion prior estimates target abundance from its typical within-genus proportion and the abundance of currently visible congeneric species. These comparisons assess the contribution of community context to recovering rare and common species.

\subsection{Representation consistency across taxonomic vocabularies (Table~\ref{tab:cross-vocabulary})}
\label{sec:appendix-vocabulary-settings}

We construct current and legacy vocabulary views for 9,044 HPP samples from the pretraining data to evaluate compositional information retention and representation consistency. Both views originate from the same abundance profile. The current view uses the MetaPhlAn 4 vocabulary; the legacy view uses identical species names and 33 curated MetaPhlAn 3--4 species mappings. TaxoMFM assigns abundance unmatched at the species level to the nearest shared genus, family, or order; abundance without a shared ancestor is removed. Structural controls include leaf-only, which retains only matched species, species-only, and random taxonomy. External baselines encode both views after mapping them to their respective model vocabularies.

Legacy abundance retention is the fraction of original abundance representable after vocabulary matching and ancestor aggregation, averaged across samples. To assess representation consistency, let $u_i$ and $v_i$ denote the L2-normalized current and legacy representations. Cosine separation measures the similarity difference between matched samples and mismatched pairs formed by a cyclic shift. For $N$ samples and shifted index $\pi(i)$,
\begin{equation}
\mathrm{Separation}=\frac{1}{N}\sum_{i=1}^{N}\left[u_i^\top v_i-u_i^\top v_{\pi(i)}\right].
\label{eq:appendix-vocabulary-separation}
\end{equation}
Bidirectional Recall@$k$ queries all samples in one view using the other view and ranks candidates by cosine similarity. It averages the fraction of correct matches retrieved among the top $k$ candidates across both directions:
\begin{equation}
\begin{aligned}
R@k=\frac{100\%}{2N}\sum_{i=1}^{N}\big[&\mathbf{1}\{\operatorname{rank}_{\mathrm{current}\to\mathrm{legacy}}(i)\le k\}\\
+&\mathbf{1}\{\operatorname{rank}_{\mathrm{legacy}\to\mathrm{current}}(i)\le k\}\big],\qquad k\in\{1,5\}.
\end{aligned}
\label{eq:appendix-vocabulary-recall}
\end{equation}
Here, $\operatorname{rank}(i)$ is the rank of the correct match for sample $i$, and $\mathbf{1}$ is the indicator function. Retention measures preserved compositional information, whereas separation and retrieval recall measure sample distinguishability; higher values are better for all metrics.

\clearpage
\section{Computational efficiency and resource requirements}
\label{sec:computational-efficiency}

We evaluated representation extraction for five models on 1,024 samples using NVIDIA RTX PRO 6000 Blackwell GPUs.
All models used existing pretrained checkpoints and BF16 mixed precision.
Measurements were repeated five times after warm-up.
Table~\ref{tab:computational-efficiency} reports steady-state inference performance, with batch size 1 for latency and batch size 8 for throughput and peak GPU memory.

\begin{table}[H]
\centering
\caption{Computational efficiency of representation extraction.}
\label{tab:computational-efficiency}
\begingroup
\definecolor{efficiencygreen}{RGB}{0,104,99}
\fontsize{8.5}{10}\selectfont
\fontencoding{OT1}\fontfamily{ptm}\selectfont
\setlength{\tabcolsep}{4pt}
\renewcommand{\arraystretch}{1.16}
\begin{tabularx}{\linewidth}{@{}>{\raggedright\arraybackslash}Xrrrrr@{}}
\toprule
\textbf{Model} &
\shortstack{\textbf{Parameters}\\\textbf{(M)}} &
\shortstack{\textbf{GFLOPs}\\\textbf{/ sample}} &
\shortstack{\textbf{Latency}\\\textbf{(ms)} $\downarrow$} &
\shortstack{\textbf{Throughput}\\\textbf{(samples/s)} $\uparrow$} &
\shortstack{\textbf{Peak memory}\\\textbf{(GiB)} $\downarrow$} \\
\midrule
MGM & 6.83 & 0.692 & 3.399 & 2,113 & 0.063 \\
Waypoint-45M & 45.15 & 4.844 & 5.732 & 1,341 & 0.237 \\
BiomeGPT & 14.73 & 1.991 & 4.262 & 1,809 & 0.114 \\
GUT-FORMer & 1.22 & 0.483 & 0.951 & 8,392 & 0.029 \\
\specialrule{0.7pt}{2pt}{1pt}
\rowcolor{efficiencygreen!8}
\textcolor{efficiencygreen}{\textbf{TaxoMFM}} & 18.25 & 5.626 & 0.759 & 3,852 & 0.412 \\
\bottomrule
\end{tabularx}
\endgroup
\end{table}

As shown in Table~\ref{tab:computational-efficiency}, TaxoMFM maintains efficient inference while incorporating multilevel taxonomic information. It achieves the lowest single-sample latency of 0.759 ms and a throughput of 3,852 samples/s, outperforming MGM, Waypoint-45M, and BiomeGPT in both measures and trailing only the lighter GUT-FORMer in throughput. Despite requiring more computation and peak memory, TaxoMFM executes efficiently on the GPU, balancing multilevel representation with inference speed.

\end{document}